\documentclass{article}

    \usepackage[final]{neurips_2025}

\usepackage[utf8]{inputenc} 
\usepackage[T1]{fontenc}    
\usepackage{hyperref}       
\usepackage{url}            
\usepackage{booktabs}       
\usepackage{amsfonts}       
\usepackage{nicefrac}       
\usepackage{microtype}      
\usepackage{xcolor}         
\usepackage{graphicx}       
\usepackage{subfig}         
\usepackage{subcaption}     
\usepackage{calc}           
\usepackage{booktabs}
\usepackage{multirow}       
\usepackage{amsmath}    
\usepackage{bbm}
\usepackage{algorithm}      
\usepackage{algpseudocode}  
\usepackage{xspace}         
\usepackage{wrapfig}        
\usepackage{kotex}          
\usepackage[disable]{todonotes} 
\usepackage{bm}
\usepackage{comment}
\usepackage{subfig}
\usepackage{amssymb}
\usepackage{url}
\usepackage{amssymb}
\usepackage{enumitem}
\usepackage[normalem]{ulem}
\usepackage{subcaption} 
\usepackage[utf8]{inputenc}
\usepackage{wrapfig}

\newcommand{\hanseul}[1]{{\color{black}#1}}

\newcommand{\ours}{ARC\xspace}

\newcommand{\seen}{In-distribution\xspace}
\newcommand{\unseen}{Zero-shot\xspace}
\newcommand{\finetune}{Few-shot Adaptation\xspace}

\definecolor{myblue}{RGB}{200,220,255}
\newcommand{\yjtodo}[1]{\todo[backgroundcolor=myblue]{\tiny #1}}

\newcommand{\ttcol}{\texttt{:}}

\newcommand{\ttA}{\texttt{A}}

\newcommand{\bbI}{\mathbb{I}}

\newcommand{\attra}{IAE\xspace}
\newcommand{\attrb}{CIE\xspace}

\title{\ours: Leveraging Compositional Representations for Cross-Problem Learning on VRPs}

\author{%
  \hanseul{Han-Seul Jeong, Youngjoon Park, Hyungseok Song, Woohyung Lim} \\
  \hanseul{LG AI Research}\\
  \hanseul{Republic of Korea} \\
  \hanseul{\texttt{ \{hanseul.jeong, yj.park, hyungseok.song, w.lim\}@lgresearch.ai } }\\
}

\begin{document}

\maketitle
\begin{abstract}

Vehicle Routing Problems (VRPs) with diverse real-world attributes have driven recent interest in cross-problem learning approaches that efficiently generalize across problem variants. We propose ARC (\textbf{A}ttribute \textbf{R}epresentation via \textbf{C}ompositional Learning), a cross-problem learning framework that learns disentangled attribute representations by decomposing them into two complementary components: an
\textit{Intrinsic Attribute Embedding (IAE)} for invariant attribute semantics and a \textit{Contextual Interaction Embedding (CIE)} for attribute-combination effects. This disentanglement is achieved by enforcing analogical consistency in the embedding space to ensure the semantic transformation of adding an attribute (e.g., a length constraint) remains invariant across different problem contexts. This enables our model to reuse invariant semantics across trained variants and construct representations for unseen combinations. 
ARC achieves state-of-the-art performance across in-distribution, zero-shot generalization, few-shot adaptation, and real-world benchmarks.

\end{abstract} 

\section{Introduction}
\label{intro}

Capacitated Vehicle Routing Problem (CVRP) represents a fundamental NP-hard combinatorial optimization challenge \citep{toth2002vehicle, symnco, garey1979computers}. While deep learning-based approximation algorithms within the Neural Combinatorial Optimization (NCO) paradigm have demonstrated near-optimal performance \citep{bengio2021machine, difusco, bqnco, lehd, am,pomo,dimes}, real-world routing applications must address diverse  attributes such as time windows \citep{solomon1987algorithms} or open routing \citep{ovrp}. To efficiently leverage information of shared attributes across multiple VRP variants, recent research has focused on cross-problem learning, where a single unified model is trained to solve multiple VRP variants defined by different attribute combinations \citep{mvmoe,routefinder,cada}, improving  efficiency and generalization compared to variant-specific models \citep{mtpomo}. 

However, prior works \citep{mtpomo, mvmoe, routefinder, cada} often conflate invariant attribute semantics with contextual effects 
among attributes, leading to entangled representations that hinder efficient knowledge sharing across different VRP variants.
To address this limitation, we propose \ours, which disentangles individual attribute embeddings by decomposing representation into intrinsic components that remain consistent across combinations and contextual components that capture combination-specific interactions.
\ours learns distinct attribute representations through analogical compositional learning, ensuring identical attributes maintain their intrinsic semantics regardless of their combinations by enforcing analogous transformations across different problem contexts.
\hanseul{Contextual components then model attribute interactions by leveraging the learned intrinsic representations within specific problem contexts, enabling efficient cross-problem learning and zero-shot generalization to unseen combinations.}

Extensive experiments demonstrate that ARC outperforms existing baselines on trained configurations while achieving robust zero-shot generalization to unseen attribute combinations and efficient few-shot adaptation to new attributes, with validation on real-world benchmarks.
Our main contributions are as followed:
\begin{itemize}[leftmargin=0.5cm, itemsep=5pt, topsep=2pt, parsep=0pt]
\item We propose ARC, a novel cross-problem learning framework that disentangles attribute representations by decomposing them into intrinsic and contextual components, facilitating effective knowledge sharing across different VRP variants.
\item We introduce a compositional learning mechanism that enforces analogical embedding relationships, establishing the first analogical embedding framework for NCO to our knowledge. 
\item We demonstrate superior performance across four scenarios: (1) in-distribution, (2) zero-shot generalization to unseen attribute combinations, (3) few-shot adaptation to new attributes, and (4) real-world benchmark, CVRPLib.
\end{itemize}
\vspace{-0.1cm}
\section{Related Works}

\textbf{Cross-Problem CO Solvers}   \hspace{0.4cm}
Recent work has shifted toward cross-problem learning, developing universal architectures capable of solving diverse problems. This research spans two branches: heterogeneous CO tasks \citep{drakulic2024goal, panunico} and VRP variants with different attribute combinations, to which our work belongs. Existing VRP approaches include joint training and Mixture-of-Experts \citep{mtpomo, mvmoe}, foundation models \citep{routefinder}, and attribute-aware attention mechanisms \citep{cada}. However, these methods learn mixed representations where shared attribute semantics are entangled with combination-specific interactions, inducing inefficient knowledge sharing across attribute combinations. Our approach explicitly decomposes attribute representations into intrinsic characteristics and interaction effects.

\textbf{Compositional Learning}  \hspace{0.4cm}
Compositional learning enables models to generalize to novel combinations by learning how individual elements can be systematically recombined. Prior approaches include modular reasoning that decomposes problems into primitive operations~\citep{johnson2017clevr, hudson2019gqa}, algebraic composition of value functions for skill reuse~\citep{van2019composing, nangue2020boolean}, and representation-level compositionality that enforces compositional structure in embedding spaces through analogy-based or contrastive objectives~\citep{arithmetic, zeroshot, chanchani2023composition}. Our approach leverages analogy-based compositionality in the embedding space for robust generalization across combinatorial tasks.

\vspace{-0.1cm}
\section{Preliminaries}\label{sec:prlim}

\subsection{Definition of VRP Variants}

Each VRP variant, including the fundamental CVRP, is defined by the constraints that correspond to the attributes activated from the set of attributes introduced below.
A CVRP instance $\bm{x}=(c_i, A_i)_{i \in \mathcal{V}}$ is defined on a complete graph $G$ with a node set $\mathcal{V} = \{0, 1, \dots, N\}$ and edge weights given by Euclidean distances $d_{ij}$, where a node $0$ represents the depot and others correspond to customers. Each node $i \in \mathcal{V}$ is associated with coordinates $c_i \in [0,1]^2$ and attribute features $A_i$ that define the constraints specific to VRP. The goal of VRP is to find an optimal solution $\bm{\tau} = (\tau_1, \dots, \tau_T)$, where $\tau_1=\tau_T=0$ and intermediate depot visits partition $\bm{\tau}$ into $K$ routes. Every customer node must be visited exactly once. The objective is to minimize the total travel distance $c(\bm{\tau}) = \sum_{t=1}^{T-1} d_{\tau_t, \tau_{t+1}}$, while satisfying all constraints defined by the attribute features $\{A_i\}_{i \in \mathcal{V}}$.

\textbf{Attribute Compositions} \hspace{0.4cm} \yjtodo{
    3) Unico 논문의 figure 1과 같은 엄청 추상화 된 그림을 여기에 넣어서 xy, constraints 구조의 VRP문제들이 서로 share 되는 부분이 된다는 것이 드러나는 그림이 있으면 읽는 사람이 '아 얘네는 VRPs를 이런식으로 이해하고 풀었구나'라고 받아들일 수 있지 않을까요? 
    링크: https://openreview.net/pdf?id=yEwakMNIex
    hssong) 그림 좋은 것 같은데 시간이 될지 모르겠네요..
}
VRP variants extend the CVRP, which includes Linehaul and Capacity ($Q$) attributes, by combining additional active attributes. Each variant must satisfy constraints from $Q$ and the active attributes.
We consider five attributes: Backhaul (B), Mixed Backhaul (MB), Open (O), Time Window (TW), and Linehaul (L). All possible VRP variants and detailed attribute specifications are provided in Appendix \ref{app:problem_variants}.
For a instance $\bm{x}$, active attributes are represented by a binary attribute indicator vector $\bbI(\bm{x})$. For attributes B, MB, O, TW, and L, this vector is $(\mathbb{I}_{\text{B}}, \mathbb{I}_{\text{MB}}, \mathbb{I}_{\text{O}}, \mathbb{I}_{\text{TW}}, \mathbb{I}_{\text{L}})$, where $\mathbb{I}_{\ttA}$ is 1 if attribute $\ttA$ is active, 0 otherwise. For example, the Open Vehicle Routing Problem with Time Windows (OVRPTW) includes O and TW, yielding $(0, 0, 1, 1, 0)$.

\subsection{Reinforcement Learning for Solving VRP Variants} 
\label{sec:RL_for_VRP}
We frame the VRP as a sequential decision-making process within a Markov Decision Process (MDP) framework, where solutions are constructed autoregressively. This approach aligns with unified modeling strategies for cross-problem learning explored in previous works \cite{routefinder, cada}.


Within the MDP formulation, at step $t$ the state $s_t=(\bm{x},\bm{\tau}_{t-1})$ consists of the instance $\bm{x}$ and the partial solution $\bm{\tau}_{t-1}=(\tau_1,\dots,\tau_{t-1})$. 
The agent selects the next node $a_t=\tau_t$ subject to feasibility, and the process starts with $\bm{\tau}_0=\emptyset$ and terminates at $t=T$ with reward $r(\bm{\tau})=-c(\bm{\tau})$.

We employ an autoregressive policy $\pi_\theta$ with an encoder $f_\theta(\bm{x})$ and a decoder $g_\theta(s_t)$. The policy defines the conditional probability $\pi_\theta(a_t | s_t) = g_\theta(\tau_t | f_\theta(\bm{x}), \bm{\tau}_{t-1})$, and the probability of generating a solution $\bm{\tau}$ is $\pi_\theta(\bm{\tau} | \bm{x}) = \prod_{t=1}^{T} \pi_\theta(a_t | s_t)$. Our goal is to maximize the expected reward $J(\theta) = \mathbb{E}_{\bm{x} \sim P(\cdot)} [\mathbb{E}_{\bm{\tau} \sim \pi_\theta(\cdot | \bm{x})} [r(\bm{\tau})]]$. 
We employ REINFORCE algorithm, augmented with the POMO \citep{pomo} and a per-variant reward normalization scheme \citep{routefinder}.
\section{Methods: Attribute Representation via Compositional learning (ARC)}

\begin{figure}[t]
  \centering
  \begin{minipage}{0.52\textwidth}
  \vspace{-0.1cm}
    \centering
\includegraphics[width=\linewidth]{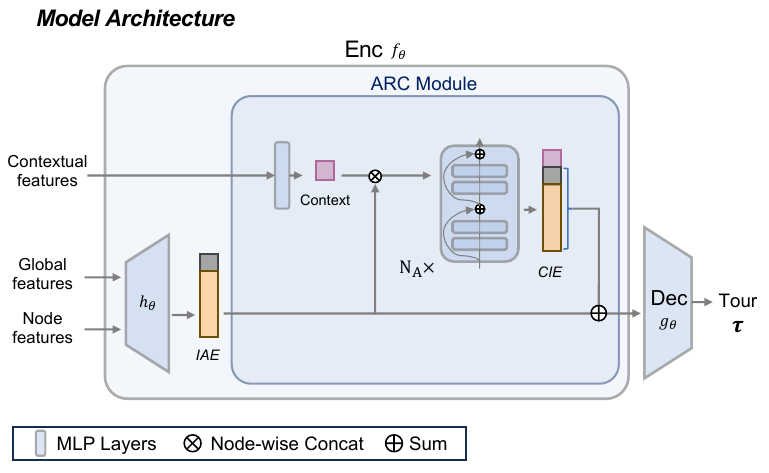}
    \caption{Overall architecture for \ours.}
    \label{fig:overall}
    \vspace{-0.1cm}
  \end{minipage}
  \hfill
  \begin{minipage}{0.47\textwidth}
  \vspace{-0.1cm}
        \centering
    \includegraphics[width=\textwidth]{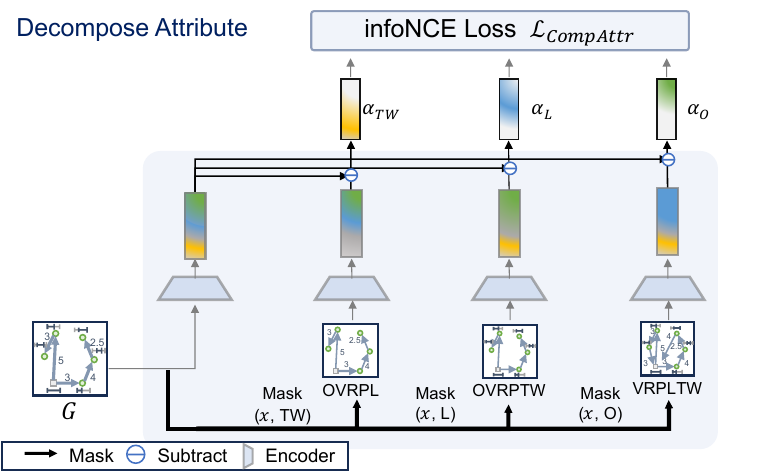}
    \caption{Compositional attribute representation learning} \label{fig:arc_comp}
  \vspace{-0.1cm}
  \end{minipage}
\end{figure}

Our approach integrates the ARC module into the encoder of a standard encoder-decoder architecture \citep{pomo}, as depicted in Figure~\ref{fig:overall}. We first identify two key properties of compositional VRPs that motivate our design (Sec. 4.1). We then detail the ARC module, which decomposes attribute representations into an \textit{Intrinsic Attribute Embedding (\attra)} for invariant semantics and a \textit{Contextual Interaction Embedding (\attrb)} for combination-specific effects (Sec. 4.2). Finally, we introduce the compositional loss designed to learn the IAE by enforcing these properties (Sec. 4.3). Full implementation details are deferred to Appendix~\ref{app:model}.


\subsection{Properties in Attributes for Compositional VRP Variants}

\emph{\textbf{P1: Intrinsic Semantics for Individual Attributes}} \hspace{0.4cm}
An attribute possesses intrinsic and invariant semantics, maintaining the same constraint definition across all attribute combinations. For example, the attribute L enforces an identical maximum route length limit in both VRPL and OVRPL, despite their distinct underlying problem structures.

\emph{\textbf{P2: Contextual Cross-Attribute Interactions}} \hspace{0.4cm}
While attributes have invariant semantics (P1), their composition yields contextual interactions beyond individual attribute effects. For instance, the influence of the attribute L is significantly attenuated when co-occurring with the open-route attribute O. This is because removing the  depot return in OVRPL substantially relaxes the length constraint, diminishing L's impact compared to its role in VRPL.
 
\subsection{Attribute Representation via Compositional Learning (ARC)}
\label{sec:attribute_mixer}


To explicitly encode properties P1 and P2, our encoder decomposes the final representation $f_{\theta}(\bm{x})$ 
into two components: an \textit{Intrinsic Attribute Embedding (IAE)}, $h_{\theta}(\bm{x})$, and a \textit{Contextual Interaction Embedding (CIE)}, $m_\theta(\bm{x})$, formulated as $f_{\theta}(\bm{x}) = h_\theta(\bm{x})+ m_\theta(\bm{x})$. The \attra is trained via a compositional loss (Sec. 4.3) to capture the intrinsic semantics of individual attributes (P1), ensuring a consistent representation for an attribute across all problem contexts. 
The \attrb, in contrast, captures the contextual interactions (P2). Conditioned on the \attra, it utilizes an attention mechanism over contextual features (i.e., attribute indicators and global features) to produce context-specific representations.
This approach allows our model to reuse the invariant semantics across variants, overcoming a key limitation of prior methods that learn mixed representations.

\subsection{Compositional Learning for Intrinsic Attribute Embeddings}
To disentangle semantic attribute representations,
we enable the model to learn analogical concepts. This section introduces the concept of analogy relationships between attribute combinations and describes our compositional loss designed to encode these relationships in the embedding space.

\textbf{Analogy-Making over Attributes}  \hspace{0.4cm}
An analogy ``A is to B as C is to D'' (denoted as A\ttcol B::C\ttcol D) captures the relationship between pairs, suggesting that the transformation from A to B parallels that from C to D.
Let $\bm{x}$ and $\bm{y}$ be two base VRP problem where an arbitrary attribute $\texttt{A}$ is not active. We denote the extended VRP variants with attribute $\texttt{A}$ activated as $\texttt{[$\bm{x} + $A]}$ and $\texttt{[$\bm{y} + $A]}$, respectively. The intrinsic semantic  of $\texttt{A}$ in both $\texttt{[$\bm{x} + $A]}$ and $\texttt{[$\bm{y} +$A]}$, as highlighted in (P1), can be expressed through the analogy: \texttt{[$\bm{x} + $A]\ttcol[$\bm{x}$]}::\texttt{[$\bm{y} + $A]\ttcol[$\bm{y}$]}.
For example, if $\bm{x}$ represents CVRP, $\bm{y}$ represents OVRP, and $\texttt{A}$ is length constraint L, this establishes the analogy \texttt{VRPL:CVRP}::\texttt{OVRPL:OVRP}.
This analogical property can be expressed in the embedding space as: 
$
    h_{\theta}(\bm{x}\text{+}\texttt{A})-h_{\theta}(\bm{x}) \approx h_{\theta}(\bm{y}\text{+}\texttt{A})-h_{\theta}(\bm{y}).$

\textbf{Learning Compositional Attribute Representation} \label{sec:infonce} \hspace{0.4cm}
To enforce the aforementioned analogical consistency on the \attra ($h_{\theta}(\bm{x})$), we employ a contrastive learning objective, InfoNCE loss \cite{infoNCE}. The key insight is that the intrinsic semantic of an attribute $\texttt{A}$ —represented by the attribute vector $\alpha_{\texttt{A}}:=h_\theta(\bm{x}+\texttt{A}) - h_\theta(\bm{x})$—should be identifiable and consistent regardless of the base instance $\bm{x}$ or other activated attributes.
It enforces this analogical consistency by distinguishing between same and different attribute semantics. As illustrated in Fig. \ref{fig:arc_comp}, we firstly extract these attribute vectors by masking the attribute features from the same instance $\bm{x}$. 
For a given attribute $\texttt{A}$, a transformation vector $h_\theta(\bm{x}'+\texttt{A}) - h_\theta(\bm{x}')$ derived from a different instance $\bm{x}'$ serves as a positive sample, while vectors corresponding to different attributes $\texttt{A}'$ form negative samples. This objective encourages the model to learn a context-invariant representation for each attribute's intrinsic semantics. This compositional loss, $\mathcal{L}_{\text{CompAttr}}(\theta)$ (formally defined in Appendix \ref{app:info}),  is added to the reward $J(\theta)$: $J(\theta)-\lambda \cdot \mathcal{L}_{\text{CompAttr}}(\theta)$.

\section{Experiments}

\textbf{Baselines} \hspace{0.4em} We compare against PyVRP \citep{pyvrp}, a state-of-the-art hybrid genetic search metaheuristic based on HGS \citep{hgs}, and recent neural cross-problem VRP solvers: MTPOMO \citep{mtpomo}, MVMoE \citep{mvmoe}, RouteFinder (RF-TE) \citep{routefinder}, and CaDA \citep{cada}. Since our method extends RouteFinder with an additional ARC module, RouteFinder can be considered an ablation of our method. Comparison with CaDA, which claims constraint-awareness, evaluates our embedding decomposition approach.

\textbf{Experimental Setup} \hspace{0.4em} Following \citep{routefinder, cada}, we train on graphs with \( N = 50 \) or \( 100 \) nodes using 100,000 instances per epoch, with equal proportions across VRP variants. We evaluate on 1,000 test instances per variant and follow RouteFinder's data generation and training protocols. The results of all neural approaches are averaged over three independent runs.
Most neural baselines share a unified codebase\footnote{\url{https://github.com/ai4co/routefinder}}, and implementations of CaDA and our method are available\hanseul{\footnote{\url{https://github.com/hanseul-jeong/ARC}}}. Detailed hyperparameters are provided in Appendix \ref{app:hyper}. We report \textbf{Gap} as the percentage cost increase relative to PyVRP's best solution and \textbf{Time} as the total duration to solve all test instances in a single run.

\begin{table*}[th!]
    \centering
    \scriptsize
    \caption{\seen Performance on 1K test instances. Bold and underline denote best and second-best, respectively. * marks the reference solution used as baseline for gap calculations.}
    \resizebox{0.8\linewidth}{!}{
    \begin{tabular}{llc@{~~}c@{~~}c@{~~}c llc@{~~}c@{~~}c@{~~}c} 
    \toprule
        & \multirow{2}{*}{Solver} & \multicolumn{2}{c}{$n = 50$} & \multicolumn{2}{c}{$n = 100$} &
        & \multirow{2}{*}{Solver} & \multicolumn{2}{c}{$n = 50$} & \multicolumn{2}{c}{$n = 100$} \\
        \cmidrule{3-6}\cmidrule{9-12}
        & & Gap (\%) & Time & Gap (\%) & Time && & Gap (\%) & Time & Gap (\%) & Time \\
        \midrule

\multirow{ 6 }{*}{\rotatebox[origin=c]{90}{CVRP}}
& HGS-PyVRP & * & 10m & * & 21m &
\multirow{6}{*}{\rotatebox[origin=c]{90}{VRPTW}}&HGS-PyVRP & * & 10m & * & 21m \\
& MTPOMO & 1.407 $\pm$ 0.01 & 1s & 2.049 $\pm$ 0.04 & 7s &&MTPOMO & 2.429 $\pm$ 0.01 & 1s & 3.934 $\pm$ 0.01 & 8s \\
& MVMoE & \underline{1.226 $\pm$ 0.00} & 2s & 1.646 $\pm$ 0.02 & 9s &&MVMoE & 2.337 $\pm$ 0.02 & 2s & 3.876 $\pm$ 0.01 & 10s \\
& RF-TE & 1.239 $\pm$ 0.05 & 1s & 1.559 $\pm$ 0.01 & 11s &&RF-TE & 1.933 $\pm$ 0.04 & 1s & 3.192 $\pm$ 0.06 & 10s \\
&  CaDA &  1.259 $\pm$ 0.05 & 3s  & \underline{ 1.513 $\pm$ 0.02} & 15s  &&CaDA & \underline{ 1.927 $\pm$ 0.07 } & 3s & \underline{ 3.025 $\pm$ 0.04 } &  16s \\
&  \ours & \textbf{ 1.154 $\pm$ 0.04} & 1s  & \textbf{ 1.429 $\pm$ 0.02} & 10s  &&\ours & \textbf{ 1.772 $\pm$ 0.07 } & 1s & \textbf{ 2.840 $\pm$ 0.04 } & 11s \\
\midrule
\multirow{ 6 }{*}{\rotatebox[origin=c]{90}{OVRP}}
& HGS-PyVRP & * & 10m & * & 21m &
\multirow{6}{*}{\rotatebox[origin=c]{90}{VRPL}}&HGS-PyVRP & * & 10m & * & 21m \\
& MTPOMO & 3.213 $\pm$ 0.02 & 1s & 5.102 $\pm$ 0.05 & 7s &&MTPOMO & 1.718 $\pm$ 0.03 & 1s & 2.486 $\pm$ 0.04 & 7s \\
& MVMoE & 2.915 $\pm$ 0.03 & 2s & 4.608 $\pm$ 0.02 & 9s &&MVMoE & 1.508 $\pm$ 0.00 & 2s & 2.065 $\pm$ 0.03 & 9s \\
& RF-TE & 2.645 $\pm$ 0.07 & 1s & 4.133 $\pm$ 0.01 & 11s &&RF-TE & \underline{1.434 $\pm$ 0.07} & 1s & 1.881 $\pm$ 0.02 & 9s \\
&  CaDA & \underline{ 2.626 $\pm$ 0.08} & 3s  & \underline{ 4.029 $\pm$ 0.01} & 15s  &&CaDA & 1.481 $\pm$ 0.07 & 3s & \underline{ 1.848 $\pm$ 0.02 } &  15s \\
&  \ours & \textbf{ 2.497 $\pm$ 0.06} & 1s  & \textbf{ 3.915 $\pm$ 0.02} & 11s  &&\ours & \textbf{ 1.370 $\pm$ 0.07 } & 1s & \textbf{ 1.753 $\pm$ 0.02 } & 11s \\
\midrule
\multirow{ 6 }{*}{\rotatebox[origin=c]{90}{VRPB}}
& HGS-PyVRP & * & 10m & * & 21m &
\multirow{6}{*}{\rotatebox[origin=c]{90}{OVRPTW}}&HGS-PyVRP & * & 10m & * & 21m \\
& MTPOMO & 3.612 $\pm$ 0.02 & 1s & 4.986 $\pm$ 0.06 & 7s &&MTPOMO & 1.564 $\pm$ 0.02 & 1s & 3.023 $\pm$ 0.03 & 8s \\
& MVMoE & 3.234 $\pm$ 0.04 & 2s & 4.484 $\pm$ 0.01 & 9s &&MVMoE & 1.528 $\pm$ 0.03 & 2s & 2.944 $\pm$ 0.06 & 10s \\
& RF-TE & 2.984 $\pm$ 0.08 & 1s & 3.999 $\pm$ 0.02 & 9s &&RF-TE & 1.285 $\pm$ 0.04 & 1s & 2.353 $\pm$ 0.07 & 11s \\
&  CaDA & \underline{ 2.973 $\pm$ 0.09} & 3s  & \underline{ 3.948 $\pm$ 0.02} & 15s  &&CaDA & \underline{ 1.240 $\pm$ 0.04 } & 3s & \underline{ 2.289 $\pm$ 0.06 } &  16s \\
&  \ours & \textbf{ 2.840 $\pm$ 0.10} & 1s  & \textbf{ 3.833 $\pm$ 0.04} & 11s  &&\ours & \textbf{ 1.087 $\pm$ 0.04 } & 1s & \textbf{ 1.968 $\pm$ 0.04 } & 12s \\
\midrule
\multirow{ 6 }{*}{\rotatebox[origin=c]{90}{VRPBL}}
& HGS-PyVRP & * & 10m & * & 21m &
\multirow{6}{*}{\rotatebox[origin=c]{90}{VRPBLTW}}&HGS-PyVRP & * & 10m & * & 21m \\
& MTPOMO & 4.681 $\pm$ 0.02 & 1s & 6.304 $\pm$ 0.04 & 7s &&MTPOMO & 2.154 $\pm$ 0.01 & 1s & 3.791 $\pm$ 0.03 & 8s \\
& MVMoE & 4.283 $\pm$ 0.03 & 2s & 5.718 $\pm$ 0.02 & 9s &&MVMoE & 2.102 $\pm$ 0.02 & 2s & 3.724 $\pm$ 0.03 & 10s \\
& RF-TE & 3.717 $\pm$ 0.11 & 1s & 5.020 $\pm$ 0.03 & 10s &&RF-TE & 1.754 $\pm$ 0.04 & 1s & 2.950 $\pm$ 0.03 & 11s \\
&  CaDA & \underline{ 3.696 $\pm$ 0.10} & 3s  & \underline{ 4.953 $\pm$ 0.03} & 15s  &&CaDA & \underline{ 1.741 $\pm$ 0.05 } & 4s & \underline{ 2.814 $\pm$ 0.05 } &  16s \\
&  \ours & \textbf{ 3.579 $\pm$ 0.13} & 1s  & \textbf{ 4.856 $\pm$ 0.01} & 11s  &&\ours & \textbf{ 1.599 $\pm$ 0.05 } & 1s & \textbf{ 2.613 $\pm$ 0.04 } & 12s \\
\midrule
\multirow{ 6 }{*}{\rotatebox[origin=c]{90}{VRPBTW}}
& HGS-PyVRP & * & 10m & * & 21m &
\multirow{6}{*}{\rotatebox[origin=c]{90}{VRPLTW}}&HGS-PyVRP & * & 10m & * & 21m \\
& MTPOMO & 1.922 $\pm$ 0.01 & 1s & 3.416 $\pm$ 0.03 & 8s &&MTPOMO & 2.857 $\pm$ 0.02 & 1s & 4.429 $\pm$ 0.01 & 8s \\
& MVMoE & 1.885 $\pm$ 0.03 & 2s & 3.360 $\pm$ 0.04 & 10s &&MVMoE & 2.747 $\pm$ 0.04 & 2s & 4.343 $\pm$ 0.03 & 10s \\
& RF-TE & 1.534 $\pm$ 0.02 & 1s & 2.597 $\pm$ 0.04 & 11s &&RF-TE & \underline{2.280 $\pm$ 0.05} & 1s & 3.620 $\pm$ 0.06 & 11s \\
&  CaDA & \underline{ 1.527 $\pm$ 0.04} & 3s  & \underline{ 2.451 $\pm$ 0.03} & 16s  &&CaDA &  2.322 $\pm$ 0.08  & 3s & \underline{ 3.453 $\pm$ 0.03 } &  16s \\
&  \ours & \textbf{ 1.375 $\pm$ 0.06} & 1s  & \textbf{ 2.281 $\pm$ 0.04} & 11s  &&\ours & \textbf{ 2.109 $\pm$ 0.06 } & 1s & \textbf{ 3.263 $\pm$ 0.05 } & 11s \\
\midrule
\multirow{ 6 }{*}{\rotatebox[origin=c]{90}{OVRPB}}
& HGS-PyVRP & * & 10m & * & 21m &
\multirow{6}{*}{\rotatebox[origin=c]{90}{OVRPBL}}&HGS-PyVRP & * & 10m & * & 21m \\
& MTPOMO & 3.000 $\pm$ 0.01 & 1s & 5.344 $\pm$ 0.03 & 7s &&MTPOMO & 3.087 $\pm$ 0.00 & 1s & 5.434 $\pm$ 0.03 & 7s \\
& MVMoE & 2.706 $\pm$ 0.02 & 2s & 4.791 $\pm$ 0.02 & 9s &&MVMoE & 2.776 $\pm$ 0.04 & 2s & 4.866 $\pm$ 0.00 & 9s \\
& RF-TE & 2.445 $\pm$ 0.10 & 1s & 4.288 $\pm$ 0.06 & 10s &&RF-TE & 2.463 $\pm$ 0.10 & 1s & 4.301 $\pm$ 0.06 & 10s \\
&  CaDA & \underline{ 2.396 $\pm$ 0.09} & 3s  & \underline{ 4.121 $\pm$ 0.03} & 15s  &&CaDA & \underline{ 2.397 $\pm$ 0.09 } & 3s & \underline{ 4.107 $\pm$ 0.02 } &  15s \\
&  \ours & \textbf{ 2.231 $\pm$ 0.10} & 1s  & \textbf{ 3.935 $\pm$ 0.04} & 11s  &&\ours & \textbf{ 2.236 $\pm$ 0.09 } & 1s & \textbf{ 3.936 $\pm$ 0.04 } & 11s \\
\midrule
\multirow{ 6 }{*}{\rotatebox[origin=c]{90}{OVRPBLTW}}
& HGS-PyVRP & * & 10m & * & 21m &
\multirow{6}{*}{\rotatebox[origin=c]{90}{OVRPBTW}}&HGS-PyVRP & * & 10m & * & 21m \\
& MTPOMO & 1.317 $\pm$ 0.01 & 1s & 2.649 $\pm$ 0.03 & 8s &&MTPOMO & 1.295 $\pm$ 0.01 & 1s & 2.622 $\pm$ 0.03 & 8s \\
& MVMoE & 1.289 $\pm$ 0.03 & 2s & 2.615 $\pm$ 0.06 & 11s &&MVMoE & 1.289 $\pm$ 0.04 & 2s & 2.604 $\pm$ 0.07 & 11s \\
& RF-TE & 1.073 $\pm$ 0.02 & 1s & 1.998 $\pm$ 0.06 & 11s &&RF-TE & 1.059 $\pm$ 0.03 & 1s & 1.997 $\pm$ 0.06 & 11s \\
&  CaDA & \underline{ 0.994 $\pm$ 0.03} & 4s  & \underline{ 1.825 $\pm$ 0.05} & 16s  &&CaDA & \underline{ 0.987 $\pm$ 0.03 } & 4s & \underline{ 1.830 $\pm$ 0.05 } &  16s \\
&  \ours & \textbf{ 0.905 $\pm$ 0.05} & 1s  & \textbf{ 1.641 $\pm$ 0.03} & 12s  &&\ours & \textbf{ 0.902 $\pm$ 0.05 } & 1s & \textbf{ 1.639 $\pm$ 0.03 } & 12s \\
\midrule
\multirow{ 6 }{*}{\rotatebox[origin=c]{90}{OVRPL}}
& HGS-PyVRP & * & 10m & * & 21m &
\multirow{6}{*}{\rotatebox[origin=c]{90}{OVRPLTW}}&HGS-PyVRP & * & 10m & * & 21m \\
& MTPOMO & 3.233 $\pm$ 0.04 & 1s & 5.154 $\pm$ 0.06 & 7s &&MTPOMO & 1.574 $\pm$ 0.01 & 1s & 3.032 $\pm$ 0.03 & 8s \\
& MVMoE & 2.966 $\pm$ 0.02 & 2s & 4.657 $\pm$ 0.02 & 9s &&MVMoE & 1.549 $\pm$ 0.03 & 2s & 2.963 $\pm$ 0.05 & 10s \\
& RF-TE & 2.636 $\pm$ 0.06 & 1s & 4.129 $\pm$ 0.02 & 10s &&RF-TE & 1.288 $\pm$ 0.04 & 1s & 2.356 $\pm$ 0.07 & 11s \\
&  CaDA & \underline{ 2.635 $\pm$ 0.09} & 3s  & \underline{ 4.029 $\pm$ 0.02} & 15s  &&CaDA & \underline{ 1.249 $\pm$ 0.05 } & 3s & \underline{ 2.290 $\pm$ 0.06 } &  16s \\
&  \ours & \textbf{ 2.510 $\pm$ 0.06} & 1s  & \textbf{ 3.916 $\pm$ 0.02} & 11s  &&\ours & \textbf{ 1.084 $\pm$ 0.04 } & 1s & \textbf{ 1.953 $\pm$ 0.04 } & 12s \\
        \bottomrule
    \label{tab:main}
    \vspace{-0.8cm}
    \end{tabular}
    }
\end{table*}

\subsection{Experimental Scenarios}
\textbf{\seen} \hspace{0.4em}
Models are trained and tested on the identical set of 16 VRP variants, encompassing all combinations of four base attributes (B, O, L, TW). This scenario evaluates the ability to capture shared attribute semantics and leverage cross-task knowledge from familiar variants.

\textbf{Out-of-distribution} \hspace{0.4em}
(1) \textbf{\unseen Generalization.} 
To evaluate performance when some attribute combinations are not provided during training, we restrict the training problem types based on the \seen setting. Models are trained on seven representative variants (CVRP, OVRP, VRPB, VRPL, VRPTW, OVRPTW, VRPBL), selected from the limited VRP variants  in \citep{mvmoe} with VRPBL added to maintain equal proportions across different attributes. We then evaluate zero-shot performance on the remaining nine variants to test compositional generalization to novel, complex attribute combinations.
(2) \textbf{\finetune.} Models pre-trained on \seen setting are extended with Efficient Adapter Layers (EAL)~\citep{routefinder} to address the unseen attribute, MB, then fine-tuned on 10,000 VRP instances with the MB attribute over 10 epochs. We compare against EAL-compatible baselines RF-TE and CaDA to assess few-shot transfer to this new constraint. (3) \textbf{Real-world Benchmark.} To validate generalization of our synthetic-trained model to real-world instances with different scales and distributions, we evaluated on 115 instances from CVRPLib\footnote{\url{http://vrp.atd-lab.inf.puc-rio.br/index.php/en/}} datasets featuring node sizes ranging from 16 to 200 and diverse distribution characteristics differing from the training data as in \citep{cada}.

\subsection{Results}
\textbf{\seen} 
\hspace{0.4em}
As shown in Table \ref{tab:main}, ARC consistently outperforms all neural baselines across all 16 VRP variants, achieving results of 1.828\% and 2.861\% for the instance sizes 50 and 100, respectively\hanseul{, without significant inference overhead and with detailed time complexity analysis provided in Appendix \ref{app:time}}. This superior performance demonstrates that ARC structurally encodes VRP characteristics at the attribute level, effectively capturing shared semantic information to interpret complex VRP variants. These experimental results provide strong empirical evidence that our proposed ARC effectively addresses the limitations of existing SOTA baselines such as MVMoE, MTPOMO, RF-TE, and CaDA, which do not utilize explicit compositional learning like ours.

 \begin{table}[t]
\centering
\begin{minipage}{0.45\textwidth}
    \centering
    
    \small
\caption{Performance on 1K test \unseen instances. (Bold: best, Underline: second-best)}

    \begin{tabular}{l@{~~~}l@{~~}c@{~~}c@{~~}} 
    \toprule
        & \multirow{1}{*}{Solver} & \multicolumn{1}{c}{$n = 50$} & \multicolumn{1}{c}{$n = 100$} \\
        \midrule
        
\multirow{5}{*}{\rotatebox[origin=c]{90}{Average}}
&MTPOMO & 5.160 $\pm$ 0.69  & 7.630 $\pm$ 1.21 \\
&MVMoE & \underline{4.613 $\pm$ 0.59} & \underline{7.524 $\pm$ 1.06} \\
&RF-TE & 5.065 $\pm$ 0.67  & 7.843 $\pm$ 0.77 \\
&CaDA & 6.740 $\pm$ 1.20  & 8.044 $\pm$ 1.39 \\
&\ours & \textbf{4.078 $\pm$ 0.42}  & \textbf{6.422 $\pm$ 0.74} \\
    \bottomrule
    \end{tabular}
    \label{tab:unseen_abs}
\end{minipage}
\hfill 
\begin{minipage}{0.52\textwidth}
    \centering
    \tiny 
    \captionof{table}{Performance on CVRPLib test instances. Bold denotes best.} 
    \label{tab:cvrplibmain}
    \begin{tabular}{lccccc}
    \toprule
    Group & MTPOMO & MVMoE & RF-TE & CaDA & \ours \\ 
    \midrule
    A & 3.23\% & 3.07\% & 2.82\% & 3.25\% & \textbf{2.50}\% \\
    B & 3.80\% & 3.89\% & 2.58\% & 2.94\% & \textbf{2.42}\% \\
    E & 8.12\% & 7.38\% & 2.93\% & 3.81\% & \textbf{2.82}\% \\
    F & 10.52\% & 12.16\% & 12.95\% & 11.96\% & \textbf{8.93}\% \\
    M & 5.61\% & 5.31\% & \textbf{5.08}\% & 5.54\% & 5.55\% \\
    P & 7.87\% & 6.76\% & 4.57\% & 5.15\% & \textbf{3.22}\% \\
    X & 5.94\% & 5.23\% & 4.48\% & 4.57\% & \textbf{4.37}\% \\
    \midrule
    Average & 6.44\% & 6.26\% & 5.06\% & 5.32\% & \textbf{4.26}\% \\
    \bottomrule
    \end{tabular}
    \label{tab:cvrplib}
\end{minipage}
\end{table}

 \begin{figure}[t]
\centering
\begin{minipage}{0.45\textwidth}
    \centering
    \includegraphics[width=\linewidth]{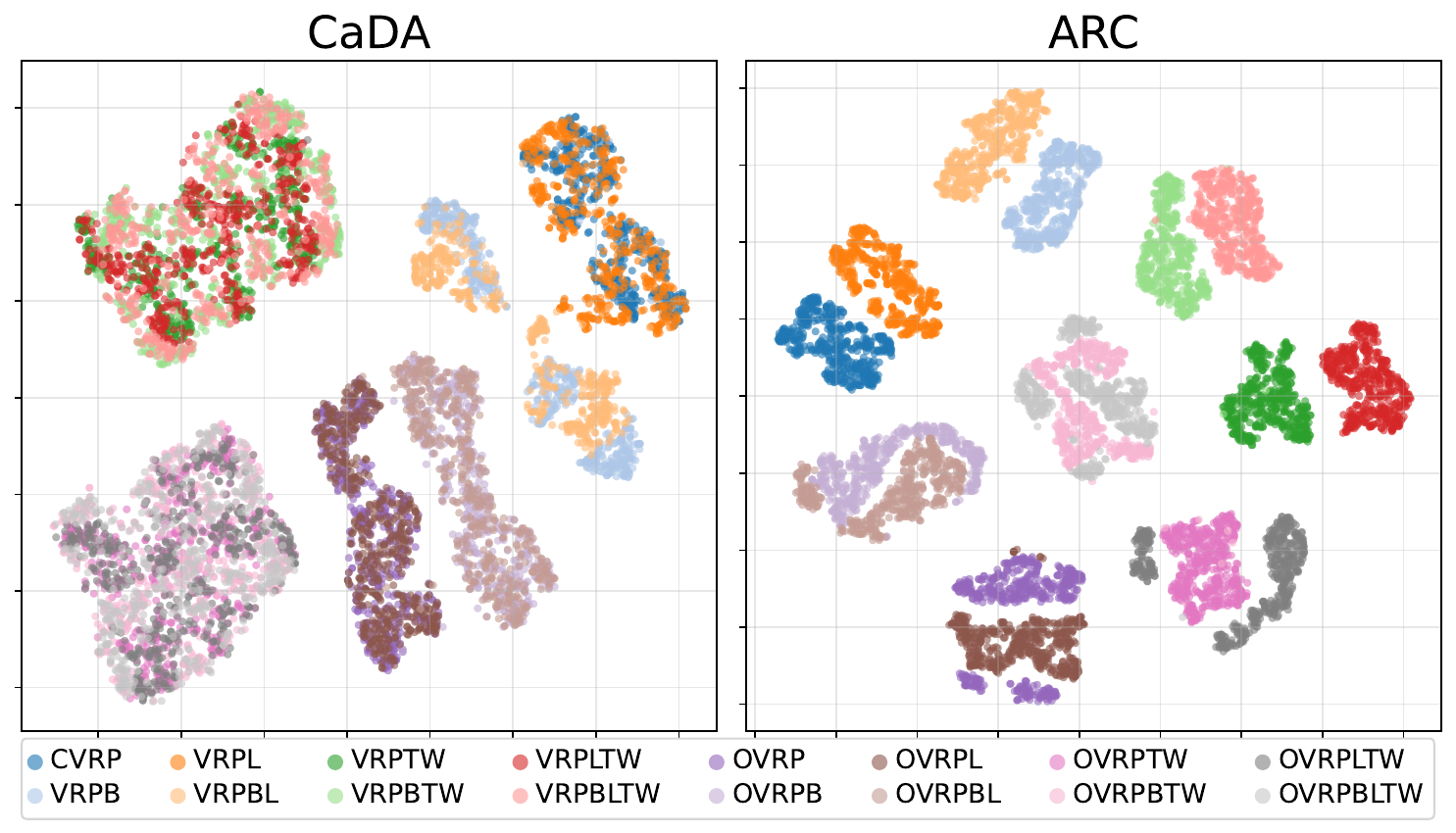}
    \caption{
        t-SNE of emb. across 16 VRP variants (\(N=50\)): CaDA (left) vs. \ours (right).
    }
    \label{fig:lossvartsne}
    \vspace{-0.2cm}
\end{minipage}
\hfill
\begin{minipage}{0.45\textwidth}
    \centering
    \includegraphics[width=1\linewidth]{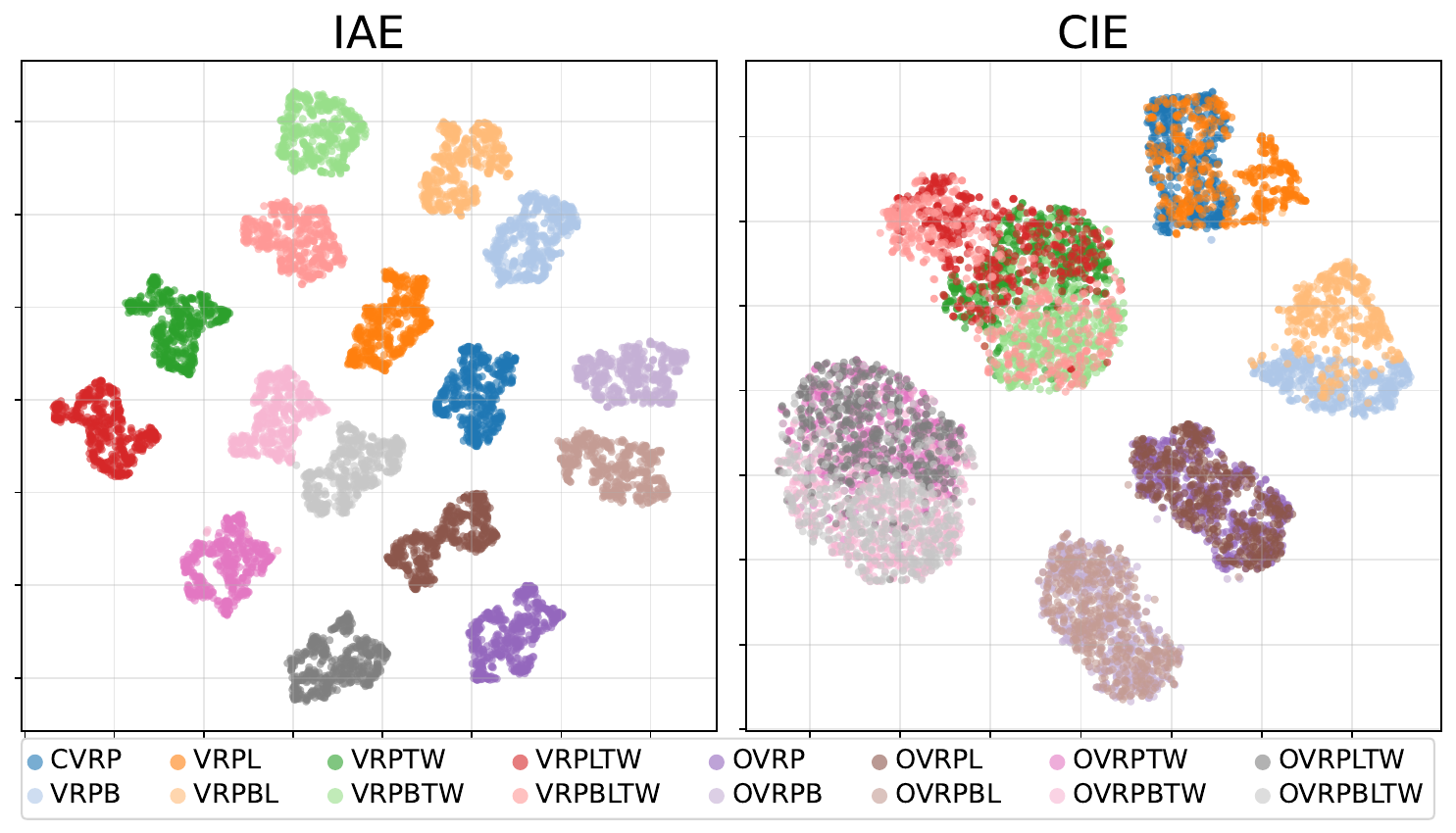}
    \caption{t-SNE of \ours emb. across 16 VRP variants (N = 50): \attra (left) vs. \attrb (right)} \label{fig:arc_comptsne_ours}
    \vspace{-0.2cm}
\end{minipage}
 \end{figure}

\textbf{Zero-shot Generalization} \hspace{0.4em}
Table \ref{tab:unseen_abs} shows the average performance across multiple tasks in \unseen settings, with complete results provided in the \hanseul{Appendix} \ref{app:experiment-result}. \ours achieves superior average performance, with MVMoE becoming the second-best algorithm compared to in-distribution results where CaDA was second-best. This shift demonstrates that compositional learning approaches (MVMoE and \ours) are more effective for generalization to unseen combinations, with \ours showing consistently strong performance across diverse tasks without large variance.

\textbf{\finetune} \hspace{0.4em}
Table \ref{tab:finetune} shows that our model consistently attains superior performance across all tasks with newly added attributes MB, significantly outperforming RF-TE and CaDA. 
Our attribute embeddings inherently preserve compatibility with unseen attribute combinations, enabling efficient reuse of learned representations for rapid adaptation to novel attributes.

\textbf{Real-World Benchmark}\label{exp:cvrplib}  \hspace{0.4cm}
We compared the test performance of \seen neural solvers trained on \(N=100\). \hanseul{As shown in Table \ref{tab:cvrplibmain}, our method achieved the best performance across all sets except Group M,
demonstrating that learning attribute-specific characteristics helps achieve superior generalization to instances with different distributional properties.
However, the reduced margins in Table \ref{tab:cvrplib-large} highlight the scaling challenge, identifying a direction for future research \citep{rethinking,mtlkd}.}

\textbf{Visualizing Embeds} \label{exp:tsne} \hspace{0.4cm}
To investigate the encoder's representation, we apply t-SNE to visualize the embeddings $f_{\theta}(\bm{x})$ of VRP instances across problem variants. Figure \ref{fig:lossvartsne} shows \ours forms well-separated clusters corresponding to respective VRP variants, while \hanseul{CaDA} shows intermixed clusters with blurred attribute boundaries. Figure \ref{fig:arc_comptsne_ours} confirms our decomposition: \attra clusters by attribute type while \attrb shows mixed patterns, validating the separation of intrinsic and contextual components.

\begin{table}[htbp]
\centering

\vspace{-0.3cm}
\tiny
    \caption{\finetune Performance on 1K test instances (\(N=50\)). Bold denotes best.} 
    \label{tab:finetune}
    \begin{tabular}{@{~~~}l@{~~~}c@{~~~}c@{~~~}c@{~~~}c@{~~~}c@{~~~}c@{~~~}c@{~~~}c@{~~~}c@{~~~}}
    \toprule
    Model & OVRPMB & OVRPMBL & OVRPMBLTW & OVRPMBTW & VRPMB & VRPMBL & VRPMBLTW & VRPMBTW \\
    \midrule
    RF-TE & $4.453 \pm 0.13$ & $4.511 \pm 0.10$ & $1.498 \pm 0.02$ & $1.486 \pm 0.02$ & $4.432 \pm 0.13$ & $3.456 \pm 0.01$ & $2.267 \pm 0.02$ & $1.945 \pm 0.02$ \\
    CaDA & $6.440 \pm 1.00$ & $6.407 \pm 0.99$ & $1.406 \pm 0.02$ & $1.391 \pm 0.02$ & $4.160 \pm 0.48$ & $4.380 \pm 0.51$ & $2.166 \pm 0.01$ & $1.853 \pm 0.01$ \\
    \ours & $\mathbf{1.475 \pm 0.06}$ & $\mathbf{1.506 \pm 0.06}$ & $\mathbf{1.277 \pm 0.04}$ & $\mathbf{1.285 \pm 0.05}$ & $\mathbf{1.677 \pm 0.05}$ & $\mathbf{1.925 \pm 0.08}$ & $\mathbf{2.062 \pm 0.03}$ & $\mathbf{1.749 \pm 0.06}$ \\
    \bottomrule
\vspace{-0.6cm}
    
\end{tabular}
\end{table}

\section{Conclusion} 


We introduced \ours, a compositional cross-problem learning framework for VRPs that disentangles attribute representations by decomposing them into intrinsic and contextual components. By enforcing analogical relationships in embedding spaces, ARC enables effective knowledge sharing across problem variants and achieves superior zero-shot generalization to unseen combinations. Extensive experiments demonstrate consistent improvements over existing baselines across in-distribution, zero-shot generalization, and few-shot adaptation, with validation on real-world benchmarks. This work establishes analogical embeddings as an effective approach for cross-problem learning in NCO. 
\hanseul{Future work will focus on extending this compositional approach to address the scalability challenges of massive-scale, realistic problem instances.}








\bibliographystyle{plain}
\bibliography{references.bib}
\clearpage
\appendix

\section{Attributes} \label{app:problem_variants}
In this section, we describe the details of attributes we utilized. 
We adopt the classical route-set notation, which is more 
convenient for presenting mathematical properties and constraints of VRP 
variants. In contrast, the main text described solutions using the sequence notation  
$\bm{\tau}=(\tau_1,\dots,\tau_T)$, where depot visits partition the sequence 
into $K$ vehicle routes, a representation well suited for reinforcement learning 
as it aligns with the step-by-step construction of solutions. 
Here, however, a solution is represented as 
$\bm{\tau}=\{\tau^1,\tau^2,\dots,\tau^K\}$, consisting of $K$ routes, where each 
route $\tau^k=(\tau^k_0,\tau^k_1,\dots,\tau^k_{n_k})$ starts and ends at the 
depot ($\tau^k_0=\tau^k_{n_k}=0$). Every customer node appears in exactly one 
route, and the total cost is $c(\bm{\tau})=\sum_{k=1}^K \sum_{i=0}^{n_k-1} d_{\tau^k_i,\tau^k_{i+1}}$. 

Based on this notation, we now detail the attributes $\{A_i\}_{i\in\mathcal{V}}$ that define different VRP variants.

\textbf{Linehaul and Capacity (Q)} \hspace{0.4cm}
In CVRP, customer nodes ($i > 0$) have a non-negative demand $q_i$, representing linehaul services (e.g., deliveries), with $A_i=\{q_i\}$. Vehicles have uniform capacity $Q>0$. The capacity attribute requires that for each route $\tau^k$, the sum of customer demands $\sum_{j \in \tau^k, j \neq 0} q_j$ must not exceed $Q$.

\textbf{Open (O)} \hspace{0.4cm} Vehicles are not required to return to the depot after serving the last customer, i.e., $\tau^k_{n_k} \neq 0$. 

\textbf{Backhaul (B) and Mixed Backhaul (MB)} \hspace{0.4cm}
Unlike the CVRP where only linehaul customers ($q_i \ge 0$) are present, Backhaul or Mixed Backhaul variants also include backhaul customers (pickup, $q_i < 0$), requiring transportation back to the depot. Each route $\tau^k$ must satisfy one of two mutually exclusive attributes: Backhaul (B), requiring all linehaul customers be visited before backhaul customers, or Mixed Backhaul (MB), allowing any order. This attribute, $\mu \in \{0, 1\}$ indicating B or MB, is a global feature included in $A_0 = \{\mu\}$.

\textbf{Time Window  (TW)} \hspace{0.4cm} Each customer node $i \neq 0$ must be visited within a time interval $[e_i, l_i] \in [0, \mathcal{T}]^2$, with a service time $s_i \in [0, \mathcal{T}]$. Customer features are thus defined as $A_i = \{e_i, l_i, s_i\}$. Vehicles arriving before the earliest available time $e_i$ must wait. Service takes $s_i$ time units, after which the vehicle proceeds. The depot has a time window $[0, \mathcal{T}]$, where $\mathcal{T}$ is the time horizon. If TW is not activated, $e_i=0, l_i=\infty, s_i=0$ are set for all customers $i$.

\textbf{Duration Limit (L)} \hspace{0.4cm} Each route’s total cost $c(\tau^{k})$ must not exceed a limit $L$, i.e., $c(\tau^{k}) \leq L$. If not activated, $L=\infty$. The global features $A_0$ include $L$, i.e., $A_0 = \{L\}$.

\begin{figure}[h]
    \centering
    \includegraphics[width=1.0\linewidth]{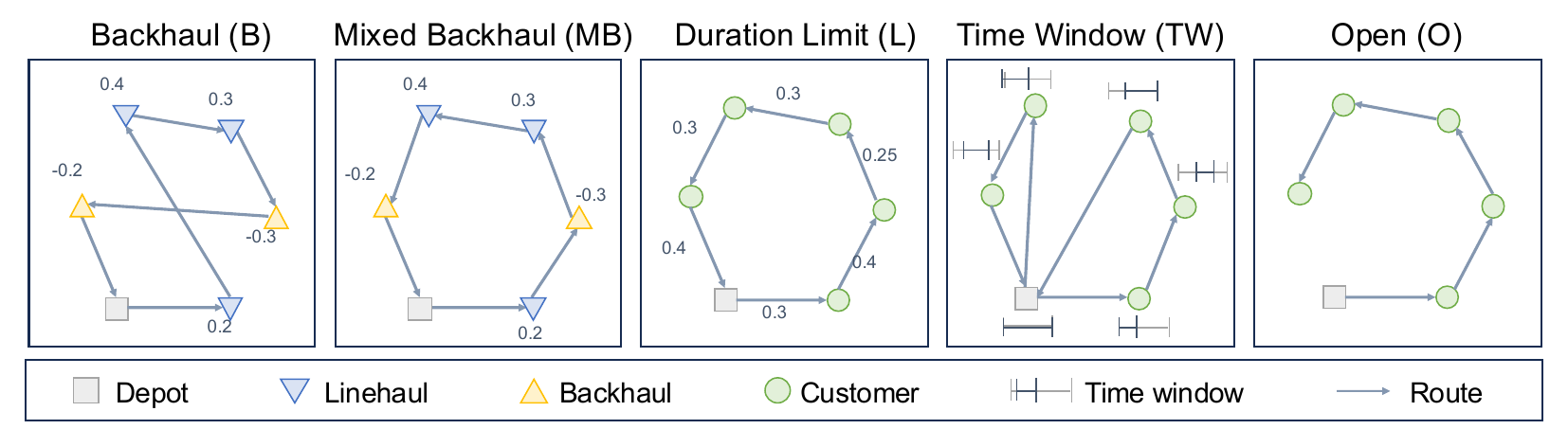}
    \caption{
llustration of VRP attributes whose combinations define respective VRP variants.
    }
    \vspace{-0.2cm}
    \label{fig:attribute}
\end{figure}

Figure \ref{fig:attribute} provides a visual explanation of the attributes. The complete set of VRP variants that can be constructed using these attributes is shown in Table \ref{tab:vrp_variants}.

\begin{table}[htbp]
\centering
\caption{VRP Variants and Their Attribute Combinations}
\label{tab:vrp_variants}
\resizebox{\textwidth}{!}{%
\begin{tabular}{l|ccccccc}
\toprule
\textbf{VRP Variant} & \textbf{Capacity} & \textbf{Open Route} & \textbf{Backhaul} & \textbf{Mixed} & \textbf{Duration Limit} & \textbf{Time Windows}\\
 & \textbf{(Q)} & \textbf{(O)} & \textbf{(B)} & \textbf{(M)} & \textbf{(L)} & \textbf{(TW)}\\
\midrule
CVRP & \checkmark & & & & & & \\
OVRP & \checkmark & \checkmark & & & & & \\
VRPB & \checkmark & & \checkmark & & & & \\
VRPL & \checkmark & & & & \checkmark & & \\
VRPTW & \checkmark & & & & & \checkmark & \\
OVRPTW & \checkmark & \checkmark & & & & \checkmark & \\
OVRPB & \checkmark & \checkmark & \checkmark & & & & \\
OVRPL & \checkmark & \checkmark & & & \checkmark & & \\
VRPBL & \checkmark & & \checkmark & & \checkmark & & \\
VRPBTW & \checkmark & & \checkmark & & & \checkmark & \\
VRPLTW & \checkmark & & & & \checkmark & \checkmark & \\
OVRPBL & \checkmark & \checkmark & \checkmark & & \checkmark & & \\
OVRPBTW & \checkmark & \checkmark & \checkmark & & & \checkmark & \\
OVRPLTW & \checkmark & \checkmark & & & \checkmark & \checkmark & \\
VRPBLTW & \checkmark & & \checkmark & & \checkmark & \checkmark & \\
OVRPBLTW & \checkmark & \checkmark & \checkmark & & \checkmark & \checkmark & \\
VRPMB & \checkmark & & \checkmark & \checkmark & & & \\
OVRPMB & \checkmark & \checkmark & \checkmark & \checkmark & & & \\
VRPMBL & \checkmark & & \checkmark & \checkmark & \checkmark & & \\
VRPMBTW & \checkmark & & \checkmark & \checkmark & & \checkmark & \\
OVRPMBL & \checkmark & \checkmark & \checkmark & \checkmark & \checkmark & & \\
OVRPMBTW & \checkmark & \checkmark & \checkmark & \checkmark & & \checkmark & \\
VRPMBLTW & \checkmark & & \checkmark & \checkmark & \checkmark & \checkmark & \\
OVRPMBLTW & \checkmark & \checkmark & \checkmark & \checkmark & \checkmark & \checkmark & \\
\bottomrule
\end{tabular}%
}
\end{table}

\section{Architecture Details} \label{app:model}

We employed a transformer-based attribute composition model, which represents a general architecture for cross-problem VRP variants \citep{mtpomo,routefinder,cada}. This approach extends single-problem models to multi-task settings by incorporating various attribute information into the structure of single models \cite{am,pomo,symnco}.

\subsection{Encoder}
The encoder $f_\theta$ comprises two principal structures: a Node Embedder $h_\theta$ and an ARC module.
The encoder separately processes two distinct categories of information: global and node-specific attributes. Global attributes including open $o \in \{0,1\}$, duration limit $d_l \in \{0,L_{\text{max}}\}$, and mixed backhaul $\mu\in\{0,1\}, $ (employed in EAL) represent problem-level constraints, whereas node-level constraints such as linehaul demand $q^l$, backhaul demand $q^b$, time window (start time $e$, end time $s$, and service time $l$) are associated with individual nodes. The global attribute information is incorporated into the depot representation, while node attribute information $\phi^n = \{\phi^n_0, \cdots, \phi^n_{H_n -1}\}$ is integrated into each customer node $i$, yielding the initial node embedding  $\bm{e}^{(0)}_{i}$ as follows:

\begin{align}
\bm{e}_i^{(0)} &= 
\begin{cases}
W_g [c_0, o, d_l, \mu, l_{0}]^T & \text{if } i = 0, \\
W_n [c_i, q^{l}_{i}, q^{b}_{i}, e_i, l_i,s_i]^T & \text{otherwise}, \\
\end{cases}
\end{align}
where $W^g \in \mathbb{R}^{\text{E}_{e} \times (\text{H}_g+2)}$ and $W^n \in \mathbb{R}^{\text{E}_{e} \times (\text{H}_n+2)}$ are learnable linear layers, $\text{E}_{e}$ is the embedding dimension, $\text{H}_{g}$ and $\text{H}_{n}$ are the numbers of global and node-level attributes respectively, and $l_{0}$ denotes the time window end value of the depot.

\subsubsection{Node Embedder}


The node embedding are passed through a NodeEmbedder $h_{\theta}$ which consists of $\text{N}_{\text{E}}$ transformer-based NodeEmbBlocks with the following structure:

\begin{align}
\bm{e}^{(\text{N}_{\text{E}})} &= h_{\theta}(\bm{e}^{(0)}) \\
&= (\text{NodeEmbBlock}_{\text{N}_{\text{E}}} \circ \text{NodeEmbBlock}_{\text{N}_{\text{E}}-1} \circ \cdots \circ \text{NodeEmbBlock}_1)(\bm{e}^{(0)}).
\end{align}


Each NodeEmbBlock consists of two sub-layers: a Multi-Head Attention (MHA) layer and a Feed Forward ParallelGatedMLP layer. The MHA layer captures dependencies between different positions in the input sequence and the ParallelGatedMLP layer applies non-linear transformations to the features. 

\begin{align}
\hat{\bm{e}}^{(\ell-1)} &= \bm{e}^{(\ell-1)} + \text{MHA}(\text{RMSNorm}(\bm{e}^{(\ell-1)}), \text{RMSNorm}(\bm{e}^{(\ell-1)})), \\
\bm{e}^{(\ell)} &= \hat{\bm{e}}^{(\ell-1)} + \text{ParallelGatedMLP}(\text{RMSNorm}(\hat{\bm{e}}^{(\ell-1)})),
\end{align}

where $\bm{e}^{(\ell-1)}$ denotes the input to the $\ell$-th NodeEmbBlock and MHA($a,b$) denotes Multi-Head Attention of which 
$a$ serves as the query, while $b$ provides the keys and values, and RMSNorm is a RMS normalization.

The ParallelGatedMLP function is defined as:

\begin{align}
\text{ParallelGatedMLP}(x) = W_{p_3}(\text{SiLU}(W_{p_1} x) \odot (W_{p_2} x)),
\end{align}

where $\odot$ denotes element-wise multiplication, SiLU is the Sigmoid Linear Unit (Swish) activation function, and $W_{p_1}$, $W_{p_2}$, $W_{p_3}$ are learnable linear layers. Consistent with prior work \citep{routefinder,cada}, we use $\text{N}_{\text{E}}=6$ layers.

\subsubsection{ARC Module}

The ARC module utilizes the final output of Node Embedder (i.e., \attra), attribute indicator, and global attribute features. Given an input $\bm{x}$ with attribute indicator ${\bbI}^{\text{attr}}$ = $(\mathbb{I}_{\text{B}}, \mathbb{I}_{\text{MB}}, \mathbb{I}_{\text{O}}, \mathbb{I}_{\text{TW}}, \mathbb{I}_{\text{L}})$, where $\mathbb{I}_{\ttA}$ is 1 if attribute $\ttA$ is active and 0 otherwise, the initial \attrb is defined as:

\begin{align}
{\textbf{m}}^{(0)} &= 
W_{m_2} \text{LayerNorm}(W_{m_1} \cdot \text{Concat}[{\bbI}^{\text{attr}},o,dl,\mu]^T ),
\end{align}

where $W_{m_1}$,  $W_{m_2}$ are learnable linear layers, $\text{Concat}$ is a feature-wise concatenation and LayerNorm is a layer normalization.

The layers of ARC module, composed of $\text{N}_\text{A}$ MixerBlocks, takes \attra $\textbf{e}$ and \attrb $\textbf{m}$ as inputs and employs a structure similar to the Global Layer \cite{cada}.
The $\ell$-th layer MixerBlock takes global embedding $\textbf{m}^{(\ell-1)}$
and \attra $\textbf{e}_{m}^{(\ell-1)}$ as inputs (i.e., $\textbf{e}_{m}^{(0)}=\textbf{e}^{(\text{N}_\text{E})}$), The MixerBlock is formulated as follows:
\begin{align}
\hat{\textbf{m}}^{(\ell-1)} &= \text{GlobalModule}(\textbf{m}^{(\ell-1)}, \textbf{e}^{(\ell-1)}_{m}), \\
\hat{\textbf{e}}^{(\ell-1)}_{m} &= \text{GlobalModule}(\textbf{e}^{(\ell-1)}_{m}, \textbf{m}^{(\ell-1)}), \\
\textbf{m}^{(\ell)} &= \hat{\textbf{m}}^{(\ell-1)} + (\hat{\textbf{e}}^{(\ell-1)}_{m}W_{g_1}), \\
\textbf{e}^{(\ell)}_{m} &= \hat{\textbf{e}}^{(\ell-1)}_{m} + (\hat{\textbf{m}}^{(\ell-1)}W_{g_2}),
\end{align}
with $W_{g_1}$ and $W_{g_2}$ are learnable linear layers. 
The GlobalModule($a$, $b$), which takes primary input $a$ and auxiliary input $b$ to produce output $\hat{a}$, is computed as follows:
\begin{align}
\tilde{a} &= \text{RMSNorm} \left( a + \text{MHA} \left( a, \text{Concat}[a, b] \right) \right), \\
\hat{a} &= \text{RMSNorm} \left( \tilde{a} + \text{ParallelGatedMLP} \left( \tilde{a} \right) \right).
\end{align}

The final output embedding of our encoder, denoted as $f_{\theta}(\bm{x})$, is computed by combining the \attra $h_{\theta}(\bm{x})$ and the \attrb $m_{\theta}(\bm{x})$. Specifically, we sum these two outputs:

\begin{equation}
    f_{\theta}(\bm{x}) = h_{\theta}(\bm{x}) +m_{\theta}(\bm{x}).
\end{equation}

\subsection{Decoder}
The decoder outputs the action probability for each node at each $t$-th step based on the encoded node embeddings. We compute the context embedding $\boldsymbol{g}_c$ using the embedding of the previously selected node $\tau_{t-1}$ and the attribute feature values at $t$-th step as follows:

\begin{equation}
    \boldsymbol{g}_c = W_d \cdot \text{Concat}[\boldsymbol{h}_{\tau_{t-1}}, c^{l}_{t}, c^{b}_{t}, z_t, l_t, o_t]^T,
\end{equation}

where $c^l_{t}, c^b_{t}$ are remaining capacity of vehicle for linehaul and backhaul, respectively and $z_t, l_t, o_t$ are current time, the remaining length of partial solution, and the indicator of the open route, respectively.

We calculate the probability of selecting each node using the context embedding as a query and the previously encoded values $\bm{h}$ as key and value. To generate feasible solutions, we mask nodes that cannot be visited at each decoding step based on activated constraints. The probability values $u_{i}$ for actions are calculated as follows:

\begin{align}
    q_c &= \text{MHA}(\bm{g}_c, \bm{h}_{0:N}), \\
    u_i &=
    \begin{cases}
    \xi \cdot \tanh \left( \frac{q_c(\bm{h}_i)^{\top}}{\sqrt{\text{E}{q}}} \right) & \text{if } i \in I_t, \\
    -\infty & \text{otherwise},
    \end{cases}
\end{align}

where $I_t$ is a feasible node set at step $t$, $\xi$ is a clipping hyperparameter, and $\text{E}_{q}$ is the dimension of query $\bm{g}_c$. 
We compute action probability by applying Softmax to probability value $u_i$.

\subsection{Learning Compositional Attribute Representation} \label{app:info}

 Given a function $\mathcal{A}(\cdot)$ that represents the set of activated attributes in a problem instance $\bm{x}$ from a batch containing instances of various problem types, we can extract an attribute vector $\alpha$ for any non-empty subset $\texttt{A} \subseteq \mathcal{A}(\bm{x})$ (where $\mathcal{A}(\bm{x}) \neq \emptyset$) as follows:

\begin{equation}
    \alpha_{\texttt{A}} = h_\theta(\bm{x}) - h_\theta(\texttt{mask}(\bm{x},\texttt{A})),
\end{equation}


where $h_\theta(\cdot)$ is the Node Embedder and $\texttt{mask}(\bm{x},\texttt{A})$ is a masking function that removes the feature of attribute \texttt{A} from instance $\bm{x}$.
We construct an attribute pool $\mathcal{P}$ by collecting all attribute vectors from all instances in the batch. For any attribute vector $\alpha$ drawn from the pool, we classify attribute vectors corresponding to the same attribute types as the positive class and those corresponding to different attribute types as the negative class. We sample one attribute vector from the positive class as the positive sample $\alpha^+$. If the size of the negative class for $\alpha$ is $B$, we compute the compositional attribute loss as follows:

\begin{equation}
    \mathcal{L}_{\text{CompAttr}}(\theta) = -\mathbb{E}_{\alpha \sim \mathcal{P}} \left[ \log \frac{\exp(f(\alpha, \alpha^+)/\beta)}{\exp(f(\alpha, \alpha^+)/\beta) + \sum_{j=1}^{B} \exp(f(\alpha, \alpha_j^-)/\beta)} \right],
\end{equation}

where $\beta$ is a temperature.
The pseudocode for this process is presented in Algorithm \ref{code:comp}.
\vspace{1cm}
\begin{algorithm}[ht!]
\caption{Compute Compositional Loss.} 
\label{code:comp}
\small
\begin{algorithmic}[1]
\Require{Batch $\{\bm{x}_1, \bm{x}_2, \ldots, \bm{x}_B\}$, temperature $\beta$, attribute function $\mathcal{A}(\cdot)$}
\State $\mathcal{P} \gets \{\}$ \Comment{Initialize attribute pool}
\For{$i = 1$ to $B$}
    \For{$\texttt{A} \in \mathcal{A}(\bm{x}_i)$}
        \State $\alpha \gets h_\theta(\bm{x}_i) - h_\theta(\texttt{mask}(\bm{x}_i, \texttt{A}))$ \Comment{Extract attribute vector}
        \State $\mathcal{P} \gets \mathcal{P} \cup \{(\alpha, \texttt{A})\}$
    \EndFor
\EndFor

\State $\mathcal{L} \gets 0$    \Comment{Initialize Compositional Loss} 
\For{$(\alpha, \texttt{A}) \in \mathcal{P}$}
    \State $\alpha^+ \sim \{\alpha' \mid (\alpha', \texttt{A}') \in \mathcal{P}, \texttt{A}' = \texttt{A}\} \setminus \{\alpha\}$  \Comment{Sample a positive attribute vector}
    \State $\mathcal{N} \gets \{\alpha' \mid (\alpha', \texttt{A}') \in \mathcal{P}, \texttt{A}' \neq \texttt{A}\}$      \Comment{Get negative class}
    \State $\mathcal{L} \gets \mathcal{L} - \log \frac{\exp(f(\alpha, \alpha^+)/\beta)}{\exp(f(\alpha, \alpha^+)/\beta) + \sum_{\alpha^- \in \mathcal{N}} \exp(f(\alpha, \alpha^-)/\beta)}$
\EndFor
\State $\mathcal{L}_\text{CompAttr} \gets \mathcal{L} / |\mathcal{P}|$ \Comment{Averaged loss} 
\State \textbf{return} $\mathcal{L}_\text{CompAttr}$ 
\end{algorithmic}
\todo[inline, backgroundcolor=yellow!30, size=scriptsize]{
    앞에 infoNCE equation 참조로 대체하고 'calucate compositional embedding loss' 정도로 표기
}
\todo[inline, backgroundcolor=yellow!30, size=scriptsize]{
    엄밀하게 수식으로 표현한건 좋지만 전체적인 절차가 한눈에 안드러오는 단점 있음. 우측에 comment으로 보완 필요. 그리고 나름 VRP에 대한 training 인데 원래의 RL loss에 대한 optimization도 표현이 될 필요는 없을까요?
}
\end{algorithm}

\clearpage

\section{Experiments} \label{app:experiment}

\subsection{Hyperparameters} \label{app:hyper}
%

\begin{figure}[h]
    \centering
    \begin{minipage}{0.33\linewidth}
    \centering
    \includegraphics[width=\linewidth]{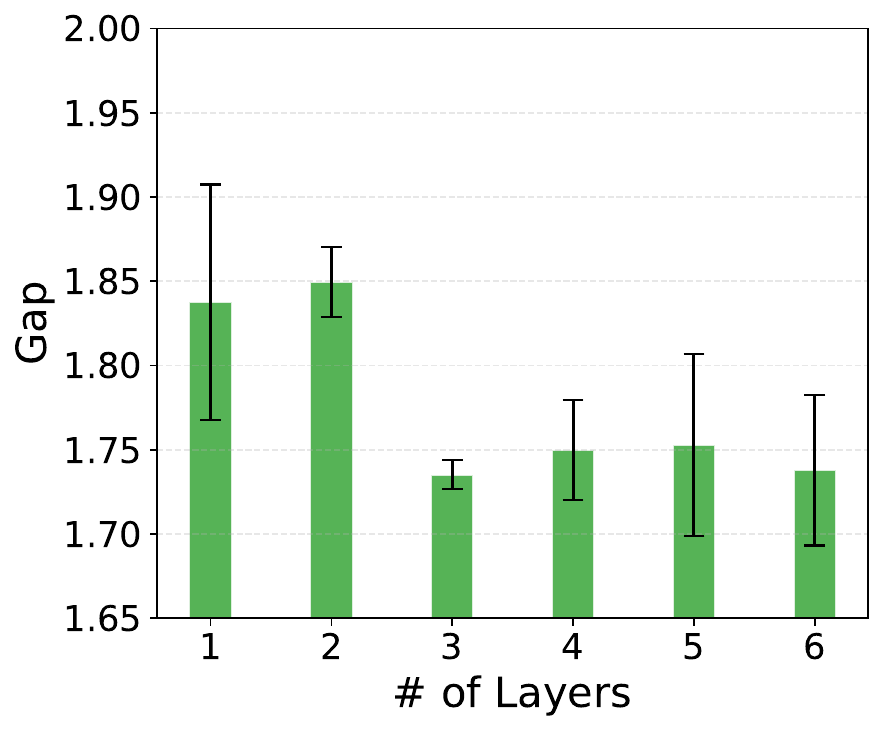}
    \caption{Params. analysis: \\ \# of Layers (ID, \(N=50\)).}
    \label{fig:layer-seen}
    \end{minipage}
        \begin{minipage}{0.33\linewidth}
    \centering
    \includegraphics[width=\linewidth]{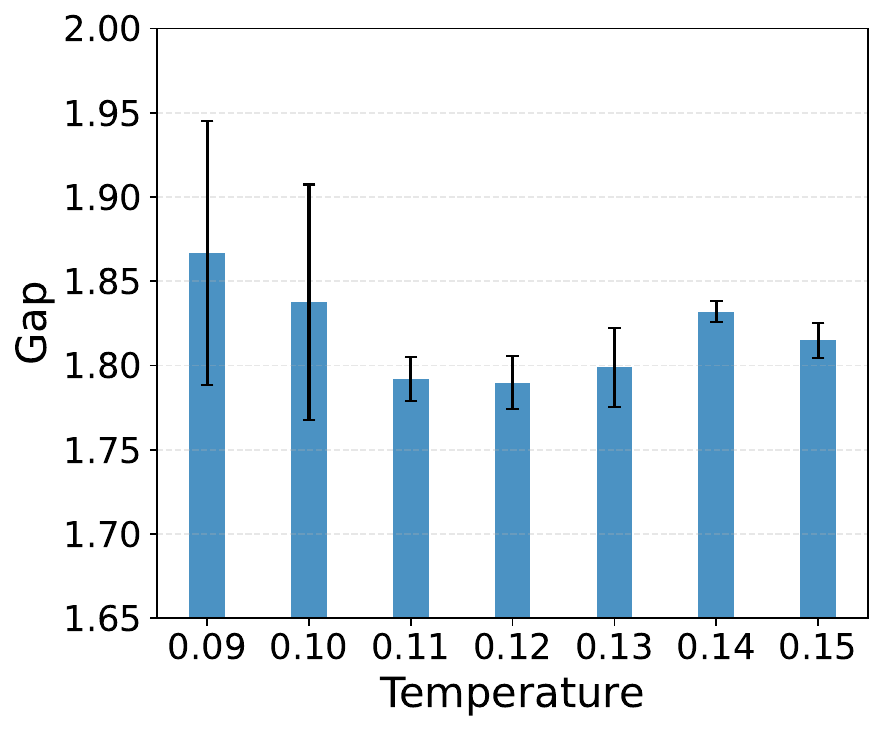}
    \caption{Params. analysis: \\ Temperature $\beta$ (ID, \(N=50\)).}
    \label{fig:temp-seen}
    \end{minipage}
        \begin{minipage}{0.32\linewidth}
    \centering
    \includegraphics[width=\linewidth]{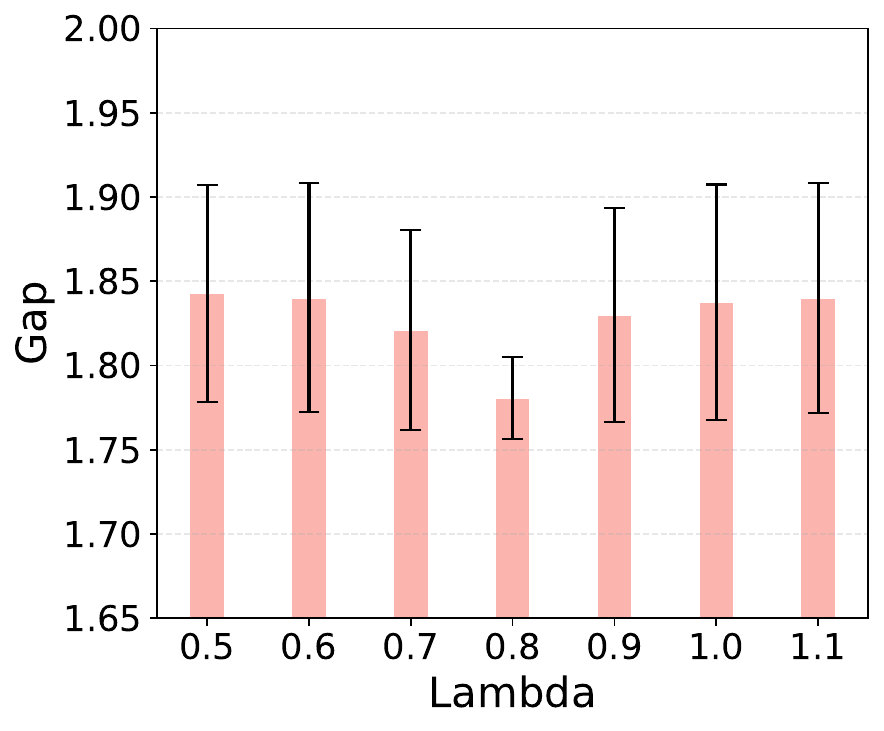}
    \caption{Params. analysis: \\ loss weight $\lambda$ (ID, \(N=50\)).}
    \label{fig:lambda-seen}
    \end{minipage}
\end{figure}

\begin{figure}[h]
    \centering
    \begin{minipage}{0.33\linewidth}
    \centering
    \includegraphics[width=\linewidth]{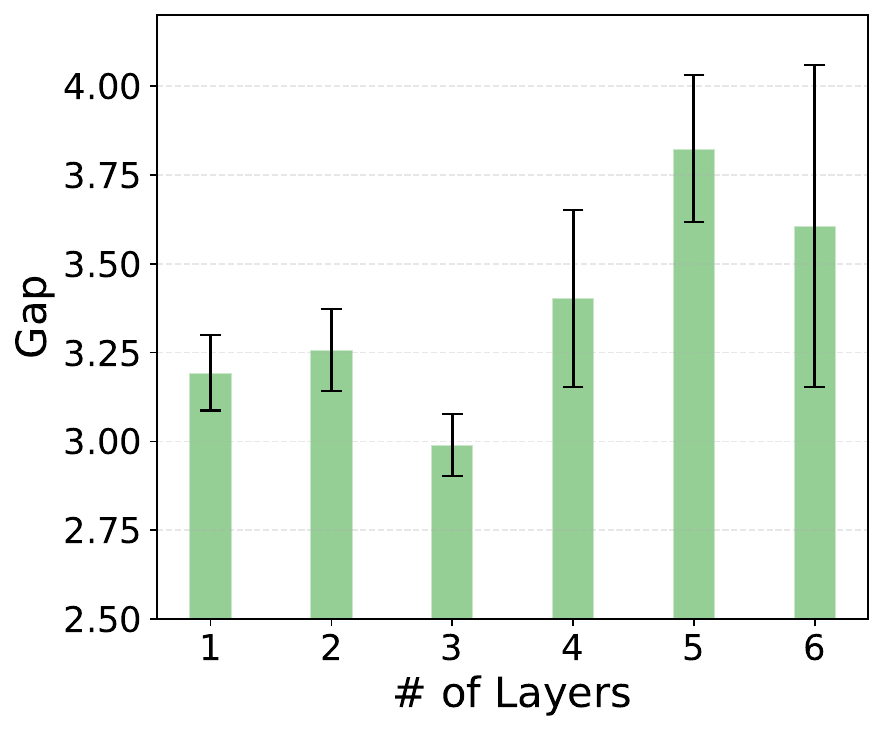}
    \caption{Params. analysis: \\ \# of Layers (OOD, \(N=50\)).}
    \label{fig:layer-unseen}
    \end{minipage}
        \begin{minipage}{0.33\linewidth}
    \centering
    \includegraphics[width=\linewidth]{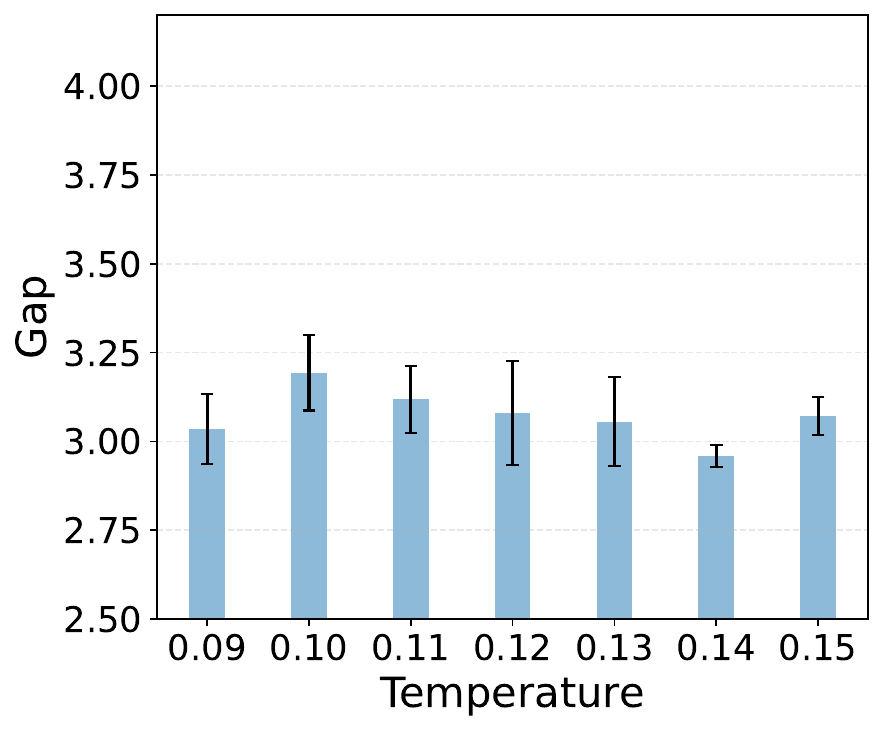}
    \caption{Params. analysis: \\ Temperature $\beta$ (OOD, \(N=50\)).}
    \label{fig:temp-unseen}
    \end{minipage}
        \begin{minipage}{0.32\linewidth}
    \centering
    \includegraphics[width=\linewidth]{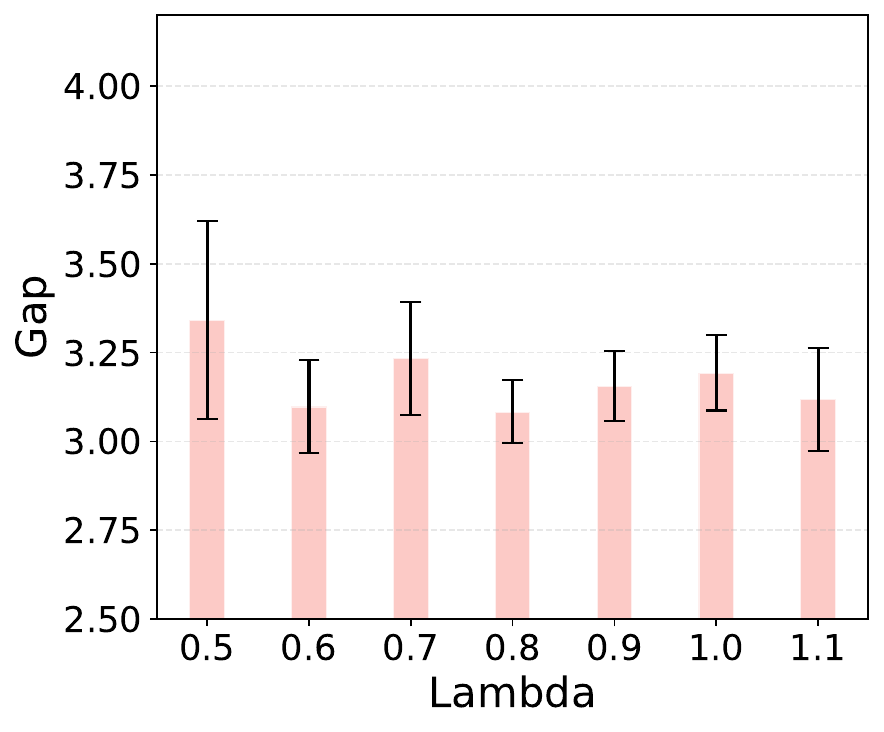}
    \caption{Params. analysis: \\ loss weight $\lambda$ (OOD, \(N=50\)).}
    \label{fig:lambda-unseen}
    \end{minipage}
\end{figure}

To assess the performance differences based on hyperparameters in our proposed \ours, we compared the average gap across 1000 samples for all problems in the validation set, using a baseline model with node size \(N=50\), the number of ARC module layers $N_A=1$, loss weight $\lambda=1.0$, temperature $\beta=0.1$. The results are presented in Figure \ref{fig:layer-seen} through Figure \ref{fig:lambda-unseen}.
Figures \ref{fig:layer-seen} and \ref{fig:layer-unseen} demonstrate performance differences based on the number of ARC module layers. Among the hyperparameters, the change in layer count $N_A$ had the most significant impact on the gap. For the \seen setting, stable and superior performance was observed when the number of layers was 3 or higher, whereas for the \unseen setting, performance significantly deteriorated with 4 or more layers. This indicates that while increasing model capacity does not reduce performance when all information is provided during training, our loss function can act as a regularizer for unseen combinations up to $N_A=3$, but cannot prevent performance degradation with further increases in model capacity.



In our contrastive loss, $\beta$ represents sensitivity to individual logit values, with larger values reducing this sensitivity. As shown in Figures \ref{fig:temp-seen} and \ref{fig:temp-unseen}, the standard error tends to decrease as $\beta$ increases, with a substantial reduction observed at values above 0.14.

However, excessively large $\beta$ values may lead to performance trade-offs by treating all logit values similarly. In our experiments, we observed that the gap decreases up to $\beta=0.12$ for the \seen setting and $\beta=0.14$ for the \unseen setting, before increasing again. Therefore, appropriate $\beta$ values should be selected based on the specific environment.

Regarding the loss scaling parameter $\lambda$, Figures \ref{fig:lambda-seen} and \ref{fig:lambda-unseen} show similar results for the \seen setting across different values, while for \unseen setting, comparable average values were observed except when selecting the very small value of 0.5. We used $N_A=3, \lambda=0.12, \lambda=0.8$, for all experiments, based on the the \seen results. Additional performance improvements may be possible through appropriate hyperparameter settings for specific experiments.

\clearpage

\subsection{Performance on Training Problems}  \label{app:experiment-result}
Table \ref{tab:unseen} shows the total performance in \unseen settings, \ours achieves superior average performance, notably 4.078\% for \(N=50\) and 6.422\% for \(N=100\). It outperforms neural baselines by an even larger margin than in \seen. This robustness is particularly evident in challenging VRP variants like OVRPB and OVRPBL. For \(N=50\), \ours significantly surpasses these baselines, achieving 4.731\% on OVRPB and 6.090\% on OVRPBL, while the baseline performance ranges from 7.121\% to 8.351\% for OVRPB and from 7.686\% to 10.984\% for OVRPBL.
This zero-shot generalization stems from \ours's explicit learning of attribute compositions, which capture the semantic structure necessary for reasoning about unseen combinations. Unlike other baselines that only implicitly model such structures, our method directly encodes attribute relationships. This demonstrates the strong generalization potential derived from a precise semantic understanding of individual attributes when encountering novel VRP variants.

\begin{table*}[th!]
    \centering
    \scriptsize
\caption{Performance on 1K test \unseen instances. Bold and underline denote best and second-best, respectively. Parentheses indicate gap changes from seen performance.}

    \begin{tabular}{l@{~~~}l@{~~}c@{~~}c@{~~}l@{~~~}l@{~~}c@{~~}c@{~~}} 
    \toprule
        & \multirow{1}{*}{Solver} & \multicolumn{1}{c}{$n = 50$} & \multicolumn{1}{c}{$n = 100$} &
        & \multirow{1}{*}{Solver} & \multicolumn{1}{c}{$n = 50$} & \multicolumn{1}{c}{$n = 100$} \\
        \midrule

\multirow{ 5 }{*}{\rotatebox[origin=c]{90}{VRPBTW}}
& MTPOMO & 4.276 $\pm$ 0.07 (+2.354) & 6.851 $\pm$ 0.10 (+3.435) &
\multirow{5}{*}{\rotatebox[origin=c]{90}{VRPLTW}}&MTPOMO & 3.172 $\pm$ 0.02 (+0.315) & 4.941 $\pm$ 0.04 (+0.512)\\
& MVMoE & \textbf{3.917 $\pm$ 0.08} (+2.032) & \underline{6.499 $\pm$ 0.00} (+3.138) &&MVMoE & 3.352 $\pm$ 0.13 (+0.604) & 5.168 $\pm$ 0.13 (+0.825)\\
& RF-TE & \underline{4.035 $\pm$ 0.05} (+2.501) & \textbf{6.240 $\pm$ 0.11} (+3.644) &&RF-TE & 3.765 $\pm$ 0.54 (+1.484) & 6.230 $\pm$ 0.32 (+2.610)\\
& CaDA & 6.067 $\pm$ 1.94 (+4.540) & 6.659 $\pm$ 0.19 (+4.208) &&CaDA & \underline{2.840 $\pm$ 0.10} (+0.519) & \underline{4.488 $\pm$ 0.15} (+1.034)\\
& \ours & 4.214 $\pm$ 0.11 (+2.839) & 6.697 $\pm$ 0.14 (+4.416) &&\ours & \textbf{2.710 $\pm$ 0.07} (+0.601) & \textbf{4.220 $\pm$ 0.03} (+0.957)\\
\midrule
\multirow{ 5 }{*}{\rotatebox[origin=c]{90}{OVRPB}}
& MTPOMO & 7.692 $\pm$ 0.26 (+4.691) & 13.735 $\pm$ 0.31 (+8.391) &
\multirow{5}{*}{\rotatebox[origin=c]{90}{OVRPBL}}&MTPOMO & 9.027 $\pm$ 0.53 (+5.940) & 13.730 $\pm$ 0.29 (+8.296)\\
& MVMoE & 7.214 $\pm$ 0.50 (+4.508) & 12.404 $\pm$ 0.08 (+7.613) &&MVMoE & \underline{7.686 $\pm$ 0.46} (+4.909) & 13.082 $\pm$ 0.22 (+8.216)\\
& RF-TE & \underline{7.121 $\pm$ 0.76} (+4.676) & 10.500 $\pm$ 0.32 (+6.212) &&RF-TE & 8.564 $\pm$ 1.33 (+6.102) & \underline{11.991 $\pm$ 1.09} (+7.690)\\
& CaDA & 8.351 $\pm$ 1.45 (+5.955) & \underline{10.224 $\pm$ 0.61} (+6.103) &&CaDA & 10.984 $\pm$ 2.08 (+8.587) & 14.435 $\pm$ 2.45 (+10.328)\\
& \ours & \textbf{4.731 $\pm$ 0.28} (+2.500) & \textbf{8.877 $\pm$ 0.35} (+4.942) &&\ours & \textbf{6.090 $\pm$ 0.19} (+3.854) & \textbf{9.694 $\pm$ 0.54} (+5.758)\\
\midrule
\multirow{ 5 }{*}{\rotatebox[origin=c]{90}{OVRPBLTW}}
& MTPOMO & 6.377 $\pm$ 1.19 (+5.061) & \underline{6.402 $\pm$ 0.50} (+3.753) &
\multirow{5}{*}{\rotatebox[origin=c]{90}{OVRPBTW}}&MTPOMO & 3.507 $\pm$ 0.07 (+2.212) & 5.960 $\pm$ 0.07 (+3.339)\\
& MVMoE & 5.031 $\pm$ 0.76 (+3.742) & 7.624 $\pm$ 0.71 (+5.009) &&MVMoE & \textbf{3.180 $\pm$ 0.08} (+1.891) & \underline{5.685 $\pm$ 0.12 }(+3.081)\\
& RF-TE & \underline{3.904 $\pm$ 0.59} (+2.831) & 7.706 $\pm$ 0.23 (+5.708) &&RF-TE & \underline{3.497 $\pm$ 0.07} (+2.438) & 5.730 $\pm$ 0.17 (+3.733)\\
& CaDA & 5.895 $\pm$ 1.00 (+4.901) & \textbf{5.949 $\pm$ 0.13} (+4.125) &&CaDA & 3.591 $\pm$ 0.10 (+2.605) & \textbf{5.673 $\pm$ 0.10} (+3.843)\\
& \ours & \textbf{3.890 $\pm$ 0.09} (+2.985) & 6.587 $\pm$ 0.11 (+4.946) &&\ours & 3.749 $\pm$ 0.10 (+2.847) & 6.334 $\pm$ 0.11 (+4.695)\\
\midrule
\multirow{ 5 }{*}{\rotatebox[origin=c]{90}{OVRPL}}
& MTPOMO & \underline{3.764 $\pm$ 0.01} (+0.531) & 5.564 $\pm$ 0.07 (+0.410) &
\multirow{5}{*}{\rotatebox[origin=c]{90}{OVRPLTW}}&MTPOMO & 3.756 $\pm$ 0.90 (+2.182) & 3.905 $\pm$ 0.10 (+0.873)\\
& MVMoE & \textbf{3.472 $\pm$ 0.02} (+0.507) & \underline{5.105 $\pm$ 0.00} (+0.448) &&MVMoE & 2.823 $\pm$ 0.49 (+1.273) & 4.403 $\pm$ 0.50 (+1.440)\\
& RF-TE & 7.142 $\pm$ 3.17 (+4.506) & 8.065 $\pm$ 1.94 (+3.936) &&RF-TE & \underline{2.791 $\pm$ 0.68 }(+1.503) & 5.152 $\pm$ 0.26 (+2.796)\\
& CaDA & 13.622 $\pm$ 4.86 (+10.987) & 14.696 $\pm$ 5.24 (+10.667) &&CaDA & 4.962 $\pm$ 1.32 (+3.714) & \underline{3.072 $\pm$ 0.18} (+0.782)\\
& \ours & 4.961 $\pm$ 0.40 (+2.451) & \textbf{4.997 $\pm$ 0.19} (+1.081) &&\ours & \textbf{1.762 $\pm$ 0.21} (+0.678) & \textbf{2.656 $\pm$ 0.16} (+0.703)\\
\midrule
\multirow{ 5 }{*}{\rotatebox[origin=c]{90}{VRPBLTW}}
& MTPOMO & 4.867 $\pm$ 0.08 (+2.713) & \underline{7.587 $\pm$ 0.18 }(+3.796) &
\multirow{5}{*}{\rotatebox[origin=c]{90}{Average}}&MTPOMO & 5.160 $\pm$ 0.69  & 7.630 $\pm$ 1.21 \\
& MVMoE & 4.846 $\pm$ 0.41 (+2.744) & 7.749 $\pm$ 0.25 (+4.024) &&MVMoE & \underline{4.613 $\pm$ 0.59} & \underline{7.524 $\pm$ 1.06} \\
& RF-TE & 4.767 $\pm$ 0.49 (+3.013) & 8.969 $\pm$ 0.58 (+6.019) &&RF-TE & 5.065 $\pm$ 0.67  & 7.843 $\pm$ 0.77 \\
& CaDA &\textbf{ 4.348 $\pm$ 0.10 }(+2.607) & \textbf{7.198 $\pm$ 0.08} (+4.385) &&CaDA & 6.740 $\pm$ 1.20  & 8.044 $\pm$ 1.39 \\
& \ours & \underline{4.591 $\pm$ 0.07} (+2.992) & {7.736 $\pm$ 0.06} (+5.123) &&\ours & \textbf{4.078 $\pm$ 0.42}  & \textbf{6.422 $\pm$ 0.74} \\

        \bottomrule
    \label{tab:unseen}
    \vspace{-0.2cm}
    \end{tabular}
\end{table*}

Table \ref{tab:unseen_seen} shows the performance of models trained with the \unseen setting on the problem variants that were included in training: CVRP, OVRP, VRPL, VRPB, VRPTW, OVRPTW, and VRPBL. Similar to Table \ref{tab:main}, \ours achieved the best performance, followed by CaDA and RF-TE. Interestingly, consistent patterns of performance improvement or deterioration compared to the \seen setting were observed across problem types, regardless of the baseline model. For VRPTW, OVRP, and OVRPTW, we can observe a general decline in neural solver performance compared to Table \ref{tab:main}.
This performance decline, despite the increased number of samples per problem type, suggests that learning from multiple problem combinations simultaneously provides greater benefits than simply increasing the sample count for individual problem types.
This suggests the effectiveness of the cross-problem approach that utilizes a unified model for learning across problem variants. Analysis of problems where our methodology showed significant improvement compared to other baselines is presented in Appendix \ref{app:gap_ratio}.

\begin{table*}[th!]
    \centering
    \scriptsize
    \caption{Performance on 1K seen problem test instances of trained with the \unseen setting. Bold and underline denote best and
second-best, respectively. Parentheses indicate gap changes from seen performance. }
    \begin{tabular}{l@{~~~}l@{~~}c@{~~}c@{~~}l@{~~~}l@{~~}c@{~~}c@{~~}} 
    \toprule
        & \multirow{1}{*}{Solver} & \multicolumn{1}{c}{$n = 50$} & \multicolumn{1}{c}{$n = 100$} &
        & \multirow{1}{*}{Solver} & \multicolumn{1}{c}{$n = 50$} & \multicolumn{1}{c}{$n = 100$} \\
        \midrule
        
\multirow{ 5 }{*}{\rotatebox[origin=c]{90}{CVRP}}
& MTPOMO & 1.197 $\pm$ 0.01 (-0.210) & 1.643 $\pm$ 0.03 (-0.406) &
\multirow{5}{*}{\rotatebox[origin=c]{90}{VRPTW}}&MTPOMO & 2.529 $\pm$ 0.03 (+0.100) & 4.198 $\pm$ 0.02 (+0.264)\\
& MVMoE & \underline{1.071 $\pm$ 0.01} (-0.155) & 1.388 $\pm$ 0.01 (-0.258) &&MVMoE & 2.494 $\pm$ 0.05 (+0.156) & 4.156 $\pm$ 0.03 (+0.281)\\
& RF-TE & 1.129 $\pm$ 0.06 (-0.110) & {1.360 $\pm$ 0.05 }(-0.200) &&RF-TE & \underline{2.094 $\pm$ 0.07} (+0.161) & 3.487 $\pm$ 0.07 (+0.295)\\
& CaDA & 1.156 $\pm$ 0.02 (-0.103) & \underline{1.306 $\pm$ 0.01} (-0.208) &&CaDA & 2.157 $\pm$ 0.09 (+0.230) & \underline{3.254 $\pm$ 0.09} (+0.229)\\
& \ours & \textbf{1.034 $\pm$ 0.04} (-0.120) & \textbf{1.225 $\pm$ 0.02} (-0.204) &&\ours & \textbf{1.864 $\pm$ 0.05} (+0.092) & \textbf{3.030 $\pm$ 0.06} (+0.190)\\
\midrule
\multirow{ 5 }{*}{\rotatebox[origin=c]{90}{OVRP}}
& MTPOMO & 3.529 $\pm$ 0.04 (+0.316) & 5.336 $\pm$ 0.07 (+0.234) &
\multirow{5}{*}{\rotatebox[origin=c]{90}{VRPL}}&MTPOMO & 1.486 $\pm$ 0.01 (-0.232) & 2.076 $\pm$ 0.02 (-0.411)\\
& MVMoE & 3.204 $\pm$ 0.00 (+0.289) & 4.883 $\pm$ 0.01 (+0.274) &&MVMoE & 1.354 $\pm$ 0.02 (-0.154) & 1.806 $\pm$ 0.02 (-0.259)\\
& RF-TE & 2.729 $\pm$ 0.09 (+0.084) & 4.157 $\pm$ 0.04 (+0.024) &&RF-TE & \underline{1.304 $\pm$ 0.08} (-0.130) & 1.655 $\pm$ 0.05 (-0.226)\\
& CaDA & \underline{2.727 $\pm$ 0.05} (+0.101) & \underline{4.027 $\pm$ 0.04 (-0.002)} &&CaDA & 1.357 $\pm$ 0.04 (-0.123) & \underline{1.604 $\pm$ 0.02} (-0.243)\\
& \ours & \textbf{2.504 $\pm$ 0.07} (+0.007) & \textbf{3.938 $\pm$ 0.05} (+0.023) &&\ours & \textbf{1.192 $\pm$ 0.05} (-0.178) & \textbf{1.526 $\pm$ 0.01} (-0.228)\\
\midrule
\multirow{ 5 }{*}{\rotatebox[origin=c]{90}{VRPB}}
& MTPOMO & 3.110 $\pm$ 0.02 (-0.501) & 4.395 $\pm$ 0.02 (-0.591) &
\multirow{5}{*}{\rotatebox[origin=c]{90}{OVRPTW}}&MTPOMO & 1.774 $\pm$ 0.01 (+0.209) & 3.394 $\pm$ 0.04 (+0.371)\\
& MVMoE & 2.855 $\pm$ 0.02 (-0.379) & 3.986 $\pm$ 0.04 (-0.498) &&MVMoE & 1.776 $\pm$ 0.04 (+0.248) & 3.342 $\pm$ 0.03 (+0.398)\\
& RF-TE & \underline{2.794 $\pm$ 0.10} (-0.191) & 3.733 $\pm$ 0.05 (-0.266) &&RF-TE & \underline{1.447 $\pm$ 0.06} (+0.162) & 2.663 $\pm$ 0.06 (+0.309)\\
& CaDA & 2.829 $\pm$ 0.07 (-0.144) & \underline{3.645 $\pm$ 0.01} (-0.303) &&CaDA & 1.485 $\pm$ 0.08 (+0.245) & \underline{2.458 $\pm$ 0.08} (+0.170)\\
& \ours & \textbf{2.556 $\pm$ 0.08} (-0.284) & \textbf{3.538 $\pm$ 0.03} (-0.295) &&\ours & \textbf{1.215 $\pm$ 0.03} (+0.128) & \textbf{2.198 $\pm$ 0.04} (+0.230)\\
\midrule

\multirow{ 5 }{*}{\rotatebox[origin=c]{90}{VRPBL}}
& MTPOMO & 4.236 $\pm$ 0.03 (-0.445) & 5.692 $\pm$ 0.03 (-0.612) &
\multirow{5}{*}{\rotatebox[origin=c]{90}{Average}}&MTPOMO & 2.551 $\pm$ 0.43 & 3.819 $\pm$ 0.58 \\
& MVMoE & 3.955 $\pm$ 0.02 (-0.328) & 5.248 $\pm$ 0.05 (-0.470) &&MVMoE & 2.387 $\pm$ 0.39  & 3.544 $\pm$ 0.56 \\
& RF-TE & \underline{3.510 $\pm$ 0.13 }(-0.206) & 4.716 $\pm$ 0.05 (-0.304) &&RF-TE & \underline{2.144 $\pm$ 0.34} & 3.110 $\pm$ 0.48 \\
& CaDa & 3.563 $\pm$ 0.04 (-0.132) & \underline{4.585 $\pm$ 0.03} (-0.368) &&CaDa & 2.182 $\pm$ 0.34  & \underline{2.983 $\pm$ 0.47} \\
& \ours & \textbf{3.185 $\pm$ 0.12} (-0.394) & \textbf{4.435 $\pm$ 0.06 }(-0.421) &&\ours & \textbf{1.936 $\pm$ 0.31}  & \textbf{2.841 $\pm$ 0.46 }\\

        \bottomrule
    \label{tab:unseen_seen}
    \end{tabular}
\end{table*}
\vspace{-10pt}

\subsection{Real-World Dataset}


To evaluate the scalability of our methodology in real-world settings, we present instance-level performance on the X group from CVRPLib.
Table \ref{tab:cvrplib-small} shows the evaluation results for instances with node sizes below 251. Similar to Table \ref{tab:cvrplibmain}, \ours demonstrated the best performance, followed by RF-TE and CaDA in order of performance.
Additionally, results for instances with node sizes above 500 are presented in Table \ref{tab:cvrplib-large}. For these larger instances, RF-TE achieved the best performance with an average gap of 12.319\%, compared to our method's 12.506\%. This suggests that while our method generalizes well to distributions different from the training set, 
it does not perform as exceptionally on very large-sized problems.
This suggests that further research on improving scalability to larger problem sizes could be beneficial.

\begin{table}[ht!]
\clearpage
\centering
\scriptsize
\caption{Performance on CVRPLib instances (\(N \leq 251\)).}
\label{tab:cvrplib-small}
\begin{tabular}{lcccccccccc}
\toprule
Set-X & \multicolumn{2}{c}{MTPOMO} & \multicolumn{2}{c}{MVMoE} & \multicolumn{2}{c}{RF-TE} & \multicolumn{2}{c}{CaDA} & \multicolumn{2}{c}{\ours} \\
Instance & Obj. & Gap & Obj. & Gap & Obj. & Gap & Obj. & Gap & Obj. & Gap \\
\midrule
X-n101-k25 & 29399 & 6.553\% & 29076 & 5.382\% & 29035 & 5.234\% & 29185 & 5.777\% & 28927 & 4.842\% \\
X-n106-k14 & 28029 & 6.323\% & 27443 & 4.101\% & 27150 & 2.989\% & 26952 & 2.238\% & 26852 & 1.859\% \\
X-n110-k13 & 15100 & 0.862\% & 15327 & 2.378\% & 15314 & 2.291\% & 15262 & 1.944\% & 15309 & 2.258\% \\
X-n115-k10 & 13412 & 5.217\% & 13475 & 5.711\% & 13338 & 4.636\% & 13169 & 3.311\% & 13458 & 5.578\% \\
X-n120-k6 & 14051 & 5.393\% & 13782 & 3.375\% & 13765 & 3.248\% & 13735 & 3.023\% & 13659 & 2.453\% \\
X-n125-k30 & 59015 & 6.259\% & 58430 & 5.205\% & 58522 & 5.371\% & 57405 & 3.360\% & 57936 & 4.316\% \\
X-n129-k18 & 30176 & 4.271\% & 29334 & 1.361\% & 29598 & 2.274\% & 29397 & 1.579\% & 29536 & 2.059\% \\
X-n134-k13 & 11707 & 7.246\% & 11462 & 5.002\% & 11585 & 6.129\% & 11512 & 5.460\% & 11605 & 6.312\% \\
X-n139-k10 & 14058 & 3.444\% & 14099 & 3.745\% & 13812 & 1.634\% & 13877 & 2.112\% & 13962 & 2.737\% \\
X-n143-k7 & 16626 & 5.898\% & 16349 & 4.134\% & 16257 & 3.548\% & 16195 & 3.153\% & 16185 & 3.089\% \\
X-n148-k46 & 46648 & 7.365\% & 45857 & 5.545\% & 45036 & 3.655\% & 45761 & 5.324\% & 45243 & 4.131\% \\
X-n153-k22 & 23514 & 10.811\% & 23649 & 11.447\% & 23478 & 10.641\% & 23154 & 9.114\% & 23299 & 9.797\% \\
X-n157-k13 & 17886 & 5.985\% & 17493 & 3.656\% & 17339 & 2.744\% & 17344 & 2.773\% & 17230 & 2.098\% \\
X-n162-k11 & 14486 & 2.461\% & 14705 & 4.010\% & 14664 & 3.720\% & 14814 & 4.781\% & 14642 & 3.565\% \\
X-n167-k10 & 21662 & 5.375\% & 21503 & 4.602\% & 21412 & 4.159\% & 21437 & 4.281\% & 21226 & 3.254\% \\
X-n172-k51 & 48560 & 6.475\% & 47883 & 4.990\% & 48118 & 5.506\% & 48181 & 5.644\% & 48022 & 5.295\% \\
X-n176-k26 & 51989 & 8.736\% & 52117 & 9.004\% & 51400 & 7.504\% & 52698 & 10.219\% & 52400 & 9.596\% \\
X-n181-k23 & 26572 & 3.923\% & 26417 & 3.317\% & 26097 & 2.065\% & 26099 & 2.073\% & 26249 & 2.659\% \\
X-n186-k15 & 25236 & 4.519\% & 25151 & 4.166\% & 25140 & 4.121\% & 25461 & 5.450\% & 25277 & 4.688\% \\
X-n190-k8 & 18222 & 7.314\% & 18988 & 11.826\% & 17892 & 5.371\% & 18470 & 8.775\% & 17877 & 5.283\% \\
X-n195-k51 & 48829 & 10.410\% & 47201 & 6.729\% & 47390 & 7.157\% & 46726 & 5.655\% & 46649 & 5.481\% \\
X-n200-k36 & 62050 & 5.927\% & 61720 & 5.364\% & 61199 & 4.474\% & 61198 & 4.473\% & 61330 & 4.698\% \\
X-n204-k19 & 20643 & 5.510\% & 20584 & 5.208\% & 20608 & 5.331\% & 20497 & 4.764\% & 20631 & 5.449\% \\
X-n209-k16 & 32298 & 5.356\% & 32358 & 5.552\% & 31876 & 3.980\% & 32092 & 4.684\% & 32170 & 4.939\% \\
X-n214-k11 & 11699 & 7.765\% & 11597 & 6.826\% & 11670 & 7.498\% & 11812 & 8.806\% & 11713 & 7.894\% \\
X-n219-k73 & 122070 & 3.805\% & 124451 & 5.830\% & 120348 & 2.341\% & 120464 & 2.440\% & 120158 & 2.180\% \\
X-n223-k34 & 43123 & 6.642\% & 42695 & 5.584\% & 42251 & 4.486\% & 42359 & 4.753\% & 42337 & 4.699\% \\
X-n228-k23 & 28233 & 9.677\% & 28171 & 9.436\% & 28798 & 11.872\% & 27988 & 8.725\% & 28098 & 9.152\% \\
X-n233-k16 & 20644 & 7.353\% & 20656 & 7.415\% & 20758 & 7.946\% & 20638 & 7.322\% & 20739 & 7.847\% \\
X-n237-k14 & 30066 & 11.183\% & 29778 & 10.118\% & 29595 & 9.441\% & 30451 & 12.606\% & 29759 & 10.047\% \\
X-n242-k48 & 88666 & 7.148\% & 87281 & 5.474\% & 85704 & 3.569\% & 85780 & 3.660\% & 85952 & 3.868\% \\
X-n247-k50 & 41610 & 11.633\% & 41345 & 10.922\% & 40639 & 9.028\% & 41037 & 10.096\% & 40817 & 9.505\% \\
X-n251-k28 & 41206 & 6.519\% & 41347 & 6.884\% & 40399 & 4.433\% & 40663 & 5.116\% & 40466 & 4.607\% \\
\midrule
Average Gap & \multicolumn{2}{c}{6.465\%} & \multicolumn{2}{c}{5.888\%} & \multicolumn{2}{c}{5.103\%} & \multicolumn{2}{c}{5.257\%} & \multicolumn{2}{c}{5.037\%} \\
\bottomrule
\end{tabular}
\end{table}

\begin{table}[ht!]
\clearpage
\centering
\scriptsize
\caption{Performance on CVRPLib instances (\(N\geq 500\)).}
\label{tab:cvrplib-large}
\begin{tabular}{lcccccccccc}
\toprule
Set-X & \multicolumn{2}{c}{MTPOMO} & \multicolumn{2}{c}{MVMoE} & \multicolumn{2}{c}{RF-TE} & \multicolumn{2}{c}{CaDA} & \multicolumn{2}{c}{\ours} \\
Instance & Obj. & Gap & Obj. & Gap & Obj. & Gap & Obj. & Gap & Obj. & Gap \\
\midrule
X-n502-k39 & 75858 & 9.580\% & 77037 & 11.283\% & 71836 & 3.770\% & 72427 & 4.624\% & 72357 & 4.523\% \\
X-n513-k21 & 34192 & 41.283\% & 32695 & 35.098\% & 28566 & 18.036\% & 30037 & 24.115\% & 29084 & 20.177\% \\
X-n524-k153 & 176706 & 14.304\% & 171622 & 11.015\% & 174075 & 12.602\% & 171656 & 11.037\% & 168443 & 8.959\% \\
X-n536-k96 & 109781 & 15.747\% & 106205 & 11.976\% & 103337 & 8.952\% & 102768 & 8.352\% & 102899 & 8.491\% \\
X-n548-k50 & 110634 & 27.606\% & 104455 & 20.479\% & 100914 & 16.394\% & 102813 & 18.585\% & 101488 & 17.057\% \\
X-n561-k42 & 55564 & 30.075\% & 53385 & 24.974\% & 49455 & 15.774\% & 50410 & 18.009\% & 49376 & 15.589\% \\
X-n573-k30 & 60460 & 19.314\% & 61611 & 21.585\% & 55937 & 10.388\% & 56622 & 11.740\% & 54785 & 8.115\% \\
X-n586-k159 & 226529 & 19.028\% & 213299 & 12.076\% & 205770 & 8.120\% & 205385 & 7.918\% & 204824 & 7.623\% \\
X-n599-k92 & 130376 & 20.217\% & 126678 & 16.807\% & 116819 & 7.716\% & 117727 & 8.553\% & 117498 & 8.342\% \\
X-n613-k62 & 78323 & 31.558\% & 73687 & 23.771\% & 67347 & 13.122\% & 68696 & 15.388\% & 68850 & 15.646\% \\
X-n627-k43 & 77282 & 24.320\% & 70710 & 13.748\% & 67339 & 8.325\% & 68838 & 10.736\% & 68295 & 9.863\% \\
X-n641-k35 & 83223 & 30.681\% & 72080 & 13.184\% & 70687 & 10.996\% & 73329 & 15.145\% & 71792 & 12.732\% \\
X-n655-k131 & 121032 & 13.347\% & 119388 & 11.807\% & 112087 & 4.970\% & 110761 & 3.728\% & 110721 & 3.691\% \\
X-n670-k130 & 182652 & 24.820\% & 166856 & 14.026\% & 169056 & 15.529\% & 165711 & 13.243\% & 162787 & 11.245\% \\
X-n685-k75 & 93216 & 36.670\% & 82525 & 20.996\% & 77687 & 13.902\% & 78145 & 14.574\% & 78560 & 15.182\% \\
X-n701-k44 & 92855 & 13.344\% & 90220 & 10.128\% & 90970 & 11.043\% & 92254 & 12.611\% & 91299 & 11.445\% \\
X-n716-k35 & 59066 & 36.181\% & 52582 & 21.232\% & 49709 & 14.608\% & 51313 & 18.306\% & 49981 & 15.235\% \\
X-n733-k159 & 175228 & 28.667\% & 156453 & 14.881\% & 148786 & 9.251\% & 148357 & 8.936\% & 147100 & 8.013\% \\
X-n749-k98 & 102540 & 32.705\% & 92308 & 19.463\% & 85048 & 10.067\% & 85634 & 10.826\% & 85627 & 10.817\% \\
X-n766-k71 & 133109 & 16.337\% & 129647 & 13.311\% & 130052 & 13.665\% & 128140 & 11.994\% & 129370 & 13.069\% \\
X-n783-k48 & 107925 & 49.097\% & 96175 & 32.864\% & 83165 & 14.891\% & 84805 & 17.157\% & 83808 & 15.779\% \\
X-n801-k40 & 92027 & 25.530\% & 87149 & 18.876\% & 86024 & 17.341\% & 89329 & 21.849\% & 87830 & 19.805\% \\
X-n819-k171 & 192568 & 21.785\% & 178857 & 13.114\% & 174609 & 10.427\% & 173461 & 9.701\% & 172161 & 8.879\% \\
X-n837-k142 & 230660 & 19.058\% & 230022 & 18.729\% & 208252 & 7.492\% & 208225 & 7.478\% & 209050 & 7.904\% \\
X-n856-k95 & 118219 & 32.883\% & 105661 & 18.767\% & 98393 & 10.597\% & 100057 & 12.468\% & 100648 & 13.132\% \\
X-n876-k59 & 114340 & 15.147\% & 114169 & 14.975\% & 107229 & 7.986\% & 110332 & 11.111\% & 108640 & 9.407\% \\
X-n895-k37 & 106400 & 97.549\% & 70002 & 29.970\% & 64525 & 19.801\% & 67511 & 25.345\% & 65160 & 20.980\% \\
X-n916-k207 & 388519 & 18.027\% & 373542 & 13.477\% & 352732 & 7.155\% & 353009 & 7.239\% & 352056 & 6.950\% \\
X-n936-k151 & 200710 & 51.234\% & 161068 & 21.364\% & 163073 & 22.875\% & 154841 & 16.672\% & 157210 & 18.457\% \\
X-n957-k87 & 126800 & 48.365\% & 123712 & 44.752\% & 102964 & 20.475\% & 105200 & 23.091\% & 103485 & 21.085\% \\
X-n979-k58 & 139389 & 17.157\% & 131894 & 10.858\% & 129770 & 9.072\% & 133144 & 11.908\% & 131469 & 10.500\% \\
X-n1001-k43 & 133680 & 84.756\% & 89126 & 23.179\% & 85998 & 18.856\% & 89231 & 23.324\% & 87914 & 21.504\% \\
\midrule
Average Gap & \multicolumn{2}{c}{30.199\%} & \multicolumn{2}{c}{18.836\%} & \multicolumn{2}{c}{12.319\%} & \multicolumn{2}{c}{13.618\%} & \multicolumn{2}{c}{12.506\%} \\
\bottomrule
\end{tabular}
\end{table}

\clearpage

\section{Analysis} \label{app:analysis}


\subsection{In-Distribution Performance: Visual Comparison of Gap Ratio}
\label{app:gap_ratio}

\begin{figure}[htbp]
    \centering
\includegraphics[width=1\linewidth]{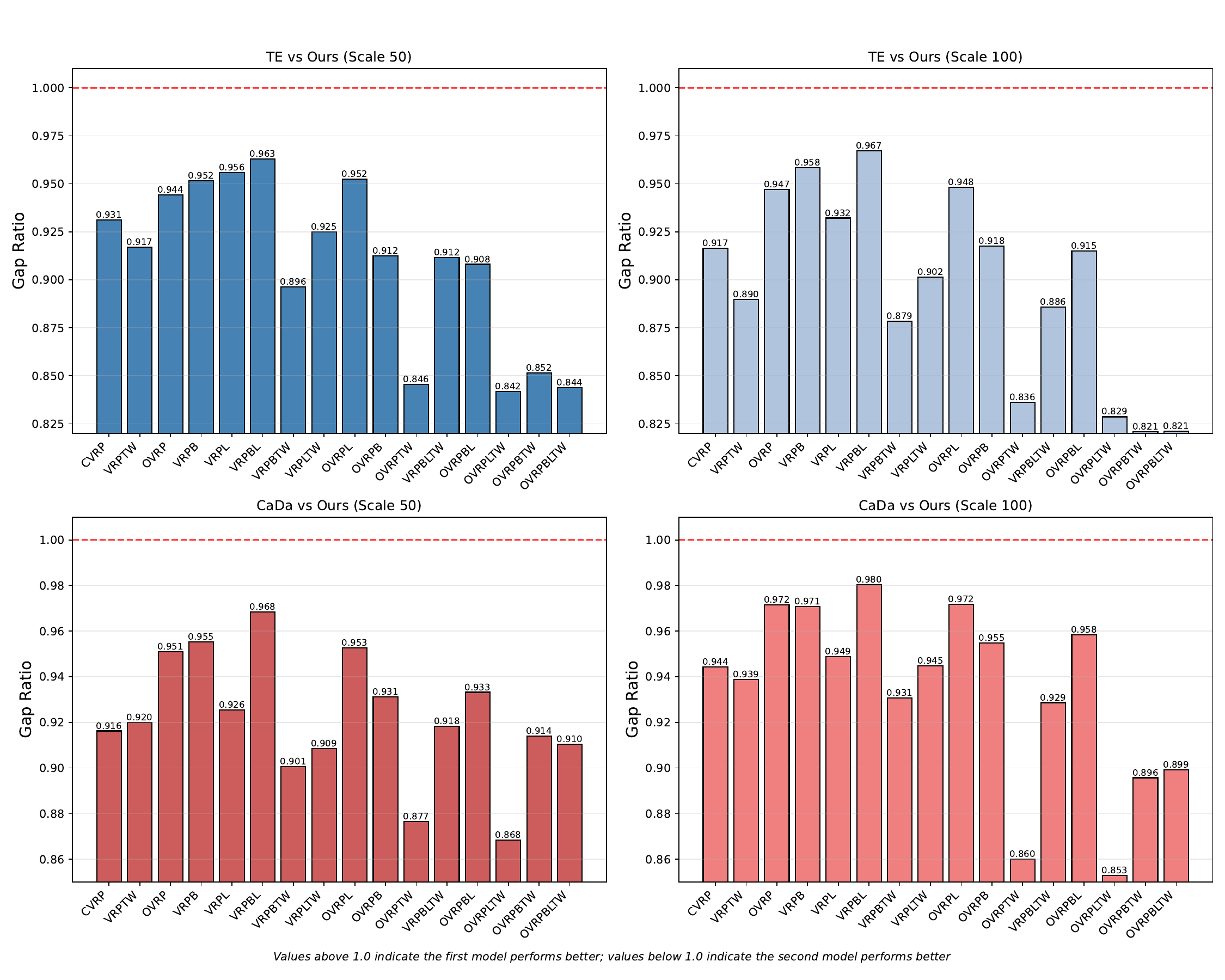}
    \caption{Visualizing Gap ratio}
    \label{fig:gap_ratio}
\end{figure}

Figure \ref{fig:gap_ratio} illustrates the gap ratios between our method \ours and comparable baselines RF-TE and CaDA, based on the performance gaps reported in Table \ref{tab:main}. The results demonstrate consistently superior performance across all problems that simultaneously incorporate both O and TW attributes (i.e., OVRPTW, OVRPLTW, OVRPBTW, OVRPBLTW). As noted in Appendix \ref{app:experiment-result}, problems such as VRPTW, OVRP, and OVRPTW showed performance improvements by leveraging information from problem instances with different combinations. This suggests that \ours achieves substantial performance gains on problems incorporating these attributes by effectively leveraging O and TW attribute information from various problem combinations through our compositional learning methodology.

\clearpage

\subsection{Train/Test Efficiency} 
\label{app:time}

\begin{figure}[htbp]
    \centering
\includegraphics[width=1\linewidth]{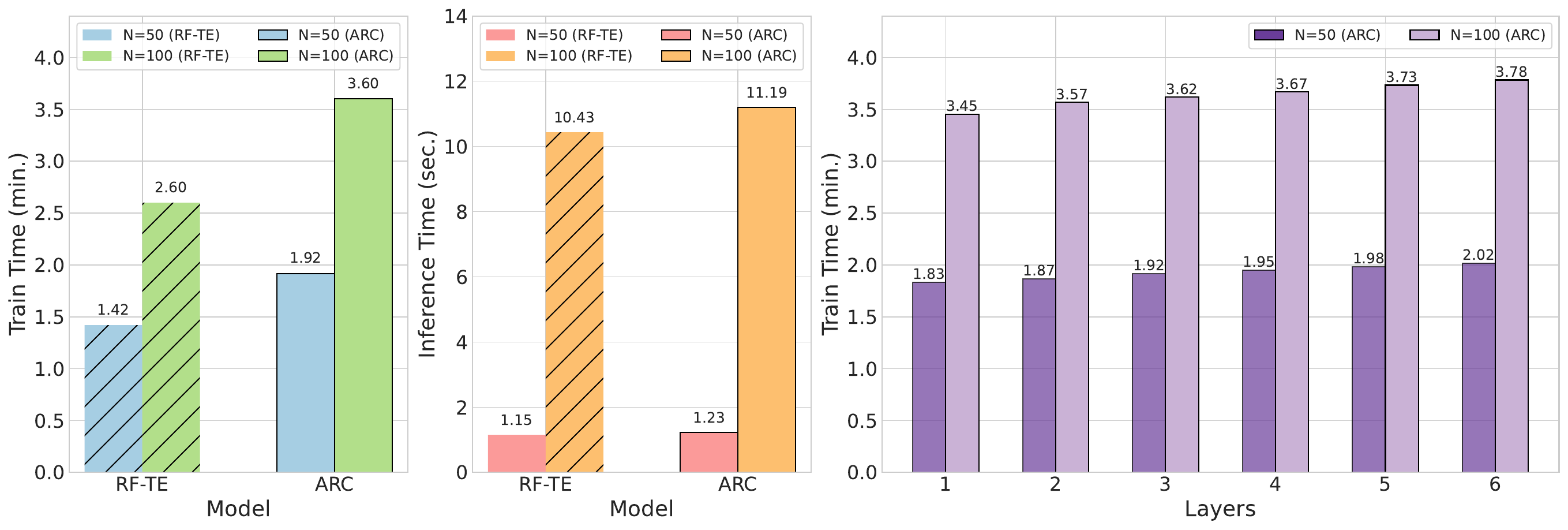}
    \caption{Computational Efficiency Comparison. Training time per epoch (left), inference time (middle), and training time by number of layers (right).}
    \label{fig:time}
\end{figure}
We compared the time required per epoch during training (100,000 instances) and evaluation (1,000 instances) for our proposed model with node sizes \(N=50\) and \(N=100\) illustrated in Figure \ref{fig:time}.
All experiments were conducted on a single
NVIDIA Tesla A40 GPU and two CPU cores of AMD EPYC 7413 24-Core Processor for 
both training and testing. For traditional solvers, we allocated 16 CPU cores
As shown in the left panel, while our method requires longer training time than RF-TE, training for \(N=100\) can be completed within a single day. Furthermore, the middle panel demonstrates that the computation of $\mathcal{L}_{\text{CompAttr}}$ is only required during training, resulting in significantly faster inference times. The right panel shows differences in training speed based on the number of ARC module layers. Due to the time-intensive nature of the sequential solution selection in the decoding step, the difference between layer counts of 1 and 6 are only 0.19 and 0.33 minutes for \(N=50\) and \(N=100\), respectively.

\end{document}